\documentclass[10pt,twocolumn,letterpaper]{article}

\makeatletter
\@namedef{ver@everyshi.sty}{}
\makeatother

\usepackage{wacv}
\usepackage{tikz}
\usepackage{times}
\usepackage{epsfig}
\usepackage{graphicx}
\usepackage{amsmath}
\usepackage{amssymb}
\usepackage{booktabs}
\usepackage{xcolor}
\usepackage{bbm}
\usepackage{tikz}

\usepackage[nocompress]{cite}

\def\wacvPaperID{126} % Enter the WACV Paper ID here

\wacvfinalcopy % *** Uncomment this line for the final submission

\ifwacvfinal
\usepackage[breaklinks=true,bookmarks=false]{hyperref}
\else
\usepackage[pagebackref=true,breaklinks=true,colorlinks,bookmarks=false]{hyperref}
\fi

\usepackage{cleveref}

\begin{document}

%%%%%%%%% TITLE
\title{Automated Detection of Label Errors in Semantic Segmentation Datasets via Deep Learning and Uncertainty Quantification}

\author{Matthias Rottmann and Marco Reese\\
University of Wuppertal\\
Dept.\ of Mathematics, IZMD\\
{\tt\small \{rottmann,reese\}@uni-wuppertal.de}
%\and
%Marco Reese\\
%University of Wuppertal\\
%Dept.\ of Mathematics, IZMD\\
%{\tt\small reese@uni-wuppertal.de}
}

\maketitle
\thispagestyle{empty}

%%%%%%%%% ABSTRACT
\begin{abstract}
    In this work, we for the first time present a method for detecting label errors in image datasets with semantic segmentation, i.e., pixel-wise class labels. Annotation acquisition for semantic segmentation datasets is time-consuming and requires plenty of human labor. In particular, review processes are time consuming and label errors can easily be overlooked by humans. The consequences are biased benchmarks and in extreme cases also performance degradation of deep neural networks (DNNs) trained on such datasets. DNNs for semantic segmentation yield pixel-wise predictions, which makes detection of label errors via uncertainty quantification a complex task. Uncertainty is particularly pronounced at the transitions between connected components of the prediction.
    By lifting the consideration of uncertainty to the level of predicted components, we enable the usage of DNNs together with component-level uncertainty quantification for the detection of label errors. We present a principled approach to benchmarking the task of label error detection by dropping labels from the Cityscapes dataset as well from a dataset extracted from the CARLA driving simulator, where in the latter case we have the labels under control. Our experiments show that our approach is able to detect the vast majority of label errors while controlling the number of false label error detections. Furthermore, we apply our method to semantic segmentation datasets frequently used by the computer vision community and present a collection of label errors along with sample statistics.
\end{abstract}

%%%%%%%%% BODY TEXT
\section{Introduction}\label{sec:introduction}

%\MR{List of inconsistencies: \\
%ground-truth vs ground truth -- go for ground truth \\
%labeling error vs label error -- go for label error \\
%dnn anstatt deep neural network \\
%segments vs components -- go for components \\
%town / towns klein schreiben \\
%attention klein \\
%CARLA vs Carla  -- go for CARLA \\
%IoU und mIoU in mathrm \\
%prefer cref over ref
%}

In many applications such as automated driving and medical imaging, large amounts of data are collected and labeled with the long-term goal of obtaining a strong predictor for such labels via artificial intelligence, in particular via deep learning \cite{DBLP:journals/corr/abs-1912-10773, Feng_2021, Hussain2019AutonomousCR, LUNDERVOLD2019102, oktay, Kamnitsas_2017}. Acquisition of so-called semantic segmentation ground truth, i.e., the pixel-wise annotation within a chosen set of classes on which we focus in this work, involves huge amounts of human labor. A German study states an effort of about 1.5 working hours per high definition street scene image \cite{FASchmidtLabeling}. Typically, industrial and scientific labelling processes consist of an iterative cycle of data labeling and quality assessment. Since the long-term goal of acquiring enough data to train e.g.\ deep neural networks (DNNs) to close to ground truth performance requires huge amounts of data, partial automation of the labeling cycle is desirable. Two research directions aiming at this goal are active learning, which aims at labeling only those data points that leverage the model performance a lot (see e.g.\ \cite{AL_SemSeg_ReBased_Bosch,colling2020metabox,settles2009active}), and the automated detection of label errors 
(see \cite{Northcutt2019,nlpannotationerror}).
%(see \cite{label_unc_Kohl2018_NeuRIPS,medical_label_Chotzoglou2019, medical_label_Ramien2019, medical_label_Zhang2020, medical_label_Tomczack2019, medical_label_disagreement_Zhang2020, label_medical_Hoebel2020, label_medical_Maier2019}). 
Currently, in active learning for semantic segmentation, a moderate number of methods exists. This is also due to the fact that active learning comes with an increased computational cost as a DNN has to be trained several times over the course of the active learning iterations \cite{AL_SemSeg_ReBased_Bosch,colling2020metabox}. Typically, these methods assume that perfect ground truth can be obtained by an oracle/teacher in each active learning iteration. In practice this is not the case and annotations are subject to multiple review loops. In that regard, current methods mostly study how noisy labels affect the model performance \cite{ImpSegLab2018,medical_label_disagreement_Zhang2020}, with the insight that DNNs can deal with a certain amount of label noise quite well. 
Methods for modeling label uncertainty in medical image segmentation, semantic street scene segmentation and everyday scene segmentation were proposed in \cite{label_unc_Kohl2018_NeuRIPS,medical_label_Tomczack2019,label_medical_Hoebel2020,medical_label_Zhang2020,Yi2021}. 

For image classification tasks, the detection of label errors was studied in \cite{Northcutt2019}. Importantly, it was pointed out that label errors harm the stability of machine learning benchmarks \cite{Northcutt2021}. This stresses the importance of being able to detect label errors, which will help to improve model benchmarks and speed up dataset review processes.

In this work, we for the first time study the task of detecting label errors in semantic segmentation in settings of low inter- and intra-observer variability. While DNNs provide predictions on pixel level, we assess DNN predictions on the level of connected components belonging to a given class by utilizing \cite{Rottmann18}. Note that this is crucial since a connected component has uncertain labels at its boundary, which makes label error detection on pixel level a complex task. For each connected component, we estimate the probability of that prediction being correct. If a connected component has a high estimated probability of being correct, while it is signaled to be false positive w.r.t.\ ground truth, we consider that component as a potential label error. We study the performance of our label error detection method on synthetic image data from the driving simulator CARLA \cite{CARLA} and on Cityscapes \cite{Cordts2016Cityscapes}. CARLA gives us a guarantee of being per se free of label errors such that we can provide a clean evaluation. To this end, we remove objects from the ground truth and study whether our method is able to identify these components as overlooked by the ground truth. Cityscapes provides high quality ground truth with only a small amount of label errors. The ground truth is available in terms of polygons such that we can drop connected components as well. In both cases it turns out that our method is able to detect most of the dropped labels while keeping the amount of false positive label errors under control. We believe that our method offers huge potential to make labeling processes more efficient. Our contribution can be summarized as follows:
\begin{itemize}
    \item We for the first time present a method that detects label errors in semantic segmentation.
    \item Utilizing \cite{Rottmann18} we detect label errors on the level of connected components.
    \item We introduce a principled benchmark for the detection of label errors based on \cite{CARLA} and \cite{Cordts2016Cityscapes}.
    \item We apply our method to additional datasets \cite{cocostuff,Everingham15,ade1} and provide examples of label errors that we found. By manually assessing samples of that data we evaluate the precision of our method on those datasets.
\end{itemize}

For all of those four real-world datasets we studied, we achieved a precision between $47.5\%$ and $67.5\%$ of correctly predicted label errors. We show that our method is able to find both overlooked and class-wise flipped labels while keeping the amount of prediction to review considerably low. %In addition, our results indicates robustness of DNNs to label noise in the dataset. 
%-------------------------------------------------------------------------

\section{Related Work}

\begin{figure*}[t]
    \center
    {\includegraphics[width=.8\textwidth]{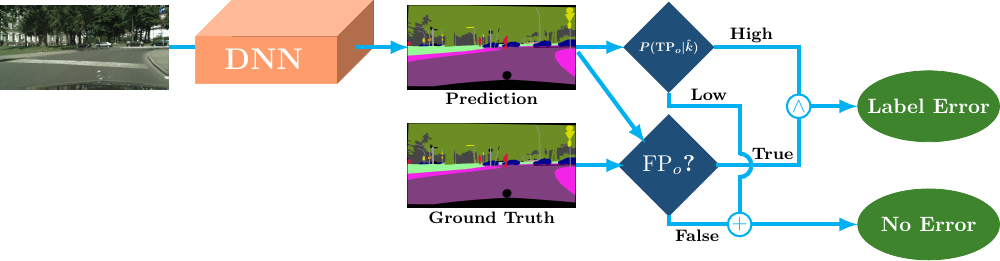}}
    \caption{A visualization of our method for label error detection.}
    \label{fig:method}
\end{figure*}

The impact of label errors on DNNs is an active field of research. Dataset labels in the context of classification have been shown to be imperfect \cite{Northcutt2021}, which also holds true for semantic segmentation. Particularly in medical images, regions of interest are often difficult to find due to visual ambiguity and inter- and intra-observer variability. For CT scans, the authors of \cite{ImpSegLab2018} observed model performance degradation when the label error noise increases. 
Hence, a number of works focus on modelling label uncertainty and develop more robust segmentation models \cite{label_unc_Kohl2018_NeuRIPS,medical_label_Tomczack2019,label_medical_Hoebel2020,medical_label_Zhang2020,Yi2021}. 
%For semantic segmentation, robust learning models are described in, e.g., \cite{ImpSegLab2018, Yi2021, label_unc_Kohl2018_NeuRIPS}.
Up to now, to the best of our knowledge the detection of incorrectly labeled ground truth connected components in semantic segmentation has not been tackled. Major challenges are the inter- and intra-observer variability in applications like medical imaging, but also the fact that segmentation networks operate on pixel level. 
%Furthermore, the typically used empirical cross-entropy loss averaged over all image pixels assumes pixel-wise independence of the model predictions, which is not true. Therefore, we cannot expect that the pixel-wise confidences will reflect the label uncertainty.

In the context of image classification, learning methods that are robust to label noise have been introduced in \cite{Northcutt2019, Hendrycks2018, Xu2019, Wang2019, Zhang2018, Jiang2018, Han2018, Goldberger2017, Reed2015}, and also the problem of label error detection has been considered in \cite{Northcutt2019}.
The authors of \cite{Northcutt2019} model class-conditional but image/instance-independent label noise in order to model label uncertainty and consider the task of label error detection. Indeed, the authors find numerous label errors in typical benchmark datasets like MNIST (image classification) or Amazon Reviews (sentiment classification). In \cite{Northcutt2021}, it is also pointed out that label errors in test sets destabilize prominent machine learning benchmarks such as MNIST, CIFAR10, CIFAR100, ImageNet, IMDB and Amazon Reviews.
Another work \cite{Chen2019} uses a cross validation approach that turns out to filter samples with noisy labels. However, for this filtering technique only the overall label quality is considered as a benchmark. In that work, label errors are not detected on an image level.
%\MR{TODO: write a bit more on the works closest to ours. Related work is not yet enough}

Our work for the first time tackles the task of label error detection in semantic segmentation, introducing a principled benchmark for the given task. An advanced post-processing method for DNN predictions lifts the problem of label error detection to the level of connected components and on that level yields calibrated estimate of the probability that a predicted connected component is indeed a correct prediction.

\section{Label Error Detection Method} \label{sec:method}

\paragraph{Estimating the probability of a prediction being correct.}

We denote a given image by $x \in [0,1]^n$ with a label $y \in \{0,\ldots,c\}^n$ being a ground truth segmentation mask over $c$ classes created by human labor. We assume to have access to a trained neural network $f$ that provides for each image pixel $z$ an estimated probability distribution $f_z(y|x)$ over the $c$ classes. Given an image $x$, let $\hat{y} = \mathrm{arg\,max}_{i} f_{\cdot}(i|x)$ denote the predicted segmentation mask and $\hat{K}=\hat{K}(x)$ the set of connected components of $\hat{y}$. Analogously, let $K=K(x)$ denote the set of connected components in the ground truth $y$. For two pixels $z,z'$ of the same predicted class, if $z'$ is in the 8-pixel neighborhood of $z$, then they both belong to the same predicted connected component $\hat{k} \in \hat{K}$. We proceed analogously for $k \in K$. %Analogously, we define the set of all connected components in the ground truth denoted by $K$.

Similarly to what is best practice in object detection, we call a connected component of the ground truth a true positive if it overlaps with a predicted connected component of the same class to a chosen degree. To this end, we use an adjusted version of the intersection over union ($\mathrm{IoU}$) from \cite{segmentme2021,Rottmann18} which is a map
$\mathrm{sIoU} : K \to [0,1]$. For $k \in K$, it is
defined as
\begin{align}\label{eq:adj_iou_gt}
\begin{split}
    &\mathrm{sIoU}(k)  := \frac{| k \cap \mathit{pr}(k)|}{  | ( k \cup \mathit{pr}( k) ) \setminus A(\hat k)|  }, \\
\text{ with} \qquad
    &\mathit{pr}(k) = \!\!\!\!\!\! \bigcup_{\hat{k} \in \hat K, \hat{k} \cap k \neq \emptyset }\!\!\! \hat k ~~~
\end{split}
\end{align}
and $A(k) = \{ z\in k': k' \in K \setminus \{k\} \}$.
The adjustment $A(k)$ excludes pixels from the union that are contained in another ground truth component $k' \in K$ of the same class, which, however, is not equal to $k$. This can happen when a predicted component intersects with multiple ground truth components. That case is punished less by the $\mathrm{sIoU}$ than by the ordinary $\mathrm{IoU}$, we refer to \cite{segmentme2021,Rottmann18} for further details.

Given a threshold $\tau\in[0,1)$, we call a ground truth component $k\in K$ true positive ($\mathrm{TP}_o$ where $o$ refers to the ``original'' task of the DNN) if $\mathrm{sIoU}(k)>\tau$, and false negative ($\mathrm{FN}_o$) otherwise.

For the remaining error, i.e., a false positive ($\mathrm{FP}_o$) connected component $\hat{k} \in \hat{K}$ of the prediction, we compute a quantity similar to the precision:
%in \Cref{eq:adj_iou_gt} %, and compute
% \begin{equation}\label{eq:adj_iou_pred}
%     \mathrm{sIoU}(\hat{k}) := \frac{|\hat{k}\cap K(\hat{k})|}{|\hat{k}\cup K(\hat{k})| - |\mathcal{A}(\hat{k})|} \, ,
% \end{equation}
%with analogous definitions of $K(\hat{k})$ and $\mathcal{A}(\hat k)$.
\begin{equation}\label{eq:prec_pred}
    \pi(\hat{k}) := \frac{|\hat{k}\cap g(\hat{k})|}{|\hat{k}|} ~~~~~\text{ with }~~~~
    \mathit{g}(\hat k) = \!\!\!\!\!\! \bigcup_{k \in K, \hat{k} \cap k \neq \emptyset }\!\!\! k \, . ~~~
\end{equation}
We call a predicted component 
% $\hat{k}\in\hat{\mathcal{K}}$ $\mathrm{FP}$ if $\mathrm{sIoU}(\hat{k})\leq\tau$. 
$\hat{k}\in\hat{K}$ $\mathrm{FP}_o$ if $\pi(\hat{k})\leq\tau$.

%We define $\hat{k} \in \hat{K}$ to be true positive (TP) if at least one pixel $z \in \hat{k}$ satisfies $y_z = \hat{y}_z$, else it is a false positive (FP) component. For $k \in K$, if none of the pixels $z \in k$ satisfy $y_z = \hat{y}_z$ we define $k$ to be a false negative (FN) ground truth component.

We utilize a so-called meta classifier that was introduced in \cite{Rottmann18}
and further extended in \cite{Rottmann19zoom,Maag20,Chan20}. It was shown empirically in \cite{colling2021false} that meta classification yields calibrated confidences on component level. Such a meta classifier computes a fixed number of $n_f$ hand-crafted features for each predicted component $\hat{k} \in \hat{K}$ yielding a structured dataset $\mathcal{M}$ of $n_f$ columns and $n_\mathit{comp} = \sum_x |\hat{K}(x)|$ rows, where $n_\mathit{comp}$ is the number of predicted components in a given number of images where the latter were not seen by the network $f$ during training (e.g.\ from a validation set). The hand-crafted features include metrics quantifying morphological features of the connected components as well as softmax uncertainty measures aggregated (average pooled) over the predicted components. For details we refer to the cited works. The meta classifier then performs a classification of $\mathrm{TP}_o$ vs.\ $\mathrm{FP}_o$ by means of the hand-crafted features. We utilize this meta classifier $m: \mathcal{M} \to [0,1]$ that yields an estimated probability for $\hat{k}$ being $\mathrm{TP}_o$, i.e., an estimate $m(\hat{k}) \approx P(\mathrm{TP}_o)$.
Due to the estimate $m(\hat{k})$ being calibrated, we can expect that it is reflective of the model's accuracy, i.e., when considering estimated probabilities in a chosen range, say 90\%--95\%, one can expect the accuracy to be in the same range.

\paragraph{Detection of label errors.}
Our label error detection method is visualized in \cref{fig:method}. It utilizes a state-of-the-art semantic segmentation DNN that is trained on a given training set. Then, we consider any sample of data not seen during training and want to detect label errors. To this end, we infer predictions from the DNN and then compare the predicted components with the ground truth components. If a component $\hat{k}$ is $\mathrm{FP}_o$, this can mean that 1) the network is indeed producing a false positive and the ground truth is likely to be correct, or 2) the ground truth is incorrect and the DNN could be correct. At this point we consider the estimate $m(\hat{k}) \approx P(\mathrm{TP}_o)$. If our calibrated meta classifier yields a estimate $m(\hat{k})$ close to $1$, the network's prediction is likely to be correct. Hence, any predicted component $\hat{k}$ being $\mathrm{FP}_o$ but having high $m(\hat{k}) \approx P(\mathrm{TP}_o)$ is considered as a candidate for review as it probably has been overlooked (or its label has been flipped) during labeling. Otherwise, we consider the prediction as not being suspicious.

Note that, in principle $P(\mathrm{TP}_o)$ can also be replaced by any uncertainty estimate that, however, must operate on the level of predicted components. On trend, the method will improve with stronger DNNs and stronger meta classifiers / uncertainty estimates, which also highlights the generic nature of our approach. As already mentioned, our approach is concerned with finding label errors in data that was not used during training. Though, our method can be easily applied to entire datasets by training in a $k$-fold cross-validation fashion.

%\MR{Meta Seg Abschnitt}

%Let $(x,y)^n \in \MR{?}$ denote a dataset of $n$ examples images $x \in \MR{?}$ with corresponding annotation $y \in \MR{?}$. In the first step of our approach, we let a neural net $f(\MR{?} | x, \omega)$ 
%predict the annotations $\tilde{y}$ for each example $x$ in the dataset. Let $K_{\tilde{y}}$ denote the set of all connected components in the predicted annotation $\tilde{y}$. We calculate the adjusted 
%IoU for every component $k \in K_{\tilde{y}}$ and for every prediction $\tilde{y}$. 
%Let $k^* \in K_{\tilde{y}}$ be a component that has an adjusted IoU below a fixed threshold $t \in [0,1]$. If the the classifier of the meta seg 
%model predicts an adjusted IoU larger than zero with a confidence above a second threshold $\tilde{t} \in [0,1]$, the component $k^*$ will be marked as a potential label error. After all segments have been 
%checked, the potential label errors will be manually verified if they are a real label error (a true positive) or not (a false positive). 

\section{Datasets and Benchmarks for Label Error Detection in Semantic Segmentation}
\label{sec:datasets}

%Not only from a technical perspective, but also on the dataset and benchmark side, the detection of label errors poses complex challenges. 
In a best case scenario, an evaluation protocol for label error detection methods has access to a perfectly clean ground truth (which does not exist in practice) where there are no label errors included, and another perturbed ground truth where there are labels missing or incorrect. A proper definition of label error itself is already difficult and subject to uncertainty.

\paragraph{Definition of label error.}
In this work, we consider a label error as a connected component of a given class that, given the context of the input image, should be contained in the ground truth, but is not contained therein.

It may happen that a spurious connected component is inserted into a ground truth, not corresponding to anything visible in the image. Such cases, however, are rare in the datasets we examined.

\paragraph{Benchmark definition.}
In our benchmark we focus on label errors that occur due to miss-outs rather than by accidentally selecting a wrong class. Note that creating such a benchmark poses a more complex task than flipping class labels. Furthermore, this is also a typical error mode that can appear due to ambiguity of visual features, size of the object and fatigue of the person performing the labeling. In our benchmark that focuses on street scenes, we remove labels from the ground truth of two chosen datasets. The labels we remove belong to the classes person, rider/bicycle, vehicles, traffic lights and road signs. In both datasets we work with high definition images and therefore drop connected components from the ground truth with pixel counts ranging $500$ and $10,\!000$. Objects with less than $500$ pixels can be considered as irrelevant, objects greater than $10000$ seem unlikely to be overlooked. We drop labels according to a Bernoulli experiment using a maximum perturbation probability $\hat{p}$ that is tied to the component size. 
More precisely, the applied perturbation probability $p$ of a component
attains its maximum of $\hat{p}$ at a component size of $500$ pixels and decreases to zero linearly until the component size reaches $10000$ pixels. This approach can be described by the equation:
\begin{equation} \label{eq:perturbation_rate}
p(k) = \mathbbm{1}_{\{ 500 \leq \lvert k \rvert \leq 10000 \}} \frac{\hat{p}(10000-|k|)}{9500},
\end{equation}
where $k \in K$ denotes a connected component from the ground truth.
%For evaluation we use a synthetic dataset created with the CARLA simulator \cite{Dosovitskiy2017CARLAAO} (version 0.9.11) and as a real street scene dataset, we use Cityscapes \cite{Cordts2016Cityscapes}. 

\paragraph{Datasets.} For our benchmark we employ a synthetic dataset generated with the CARLA driving simulator \cite{Dosovitskiy2017CARLAAO} (version 0.9.11) and the real world dataset Cityscapes \cite{Cordts2016Cityscapes}. 
For CARLA we have the ground truth under control in a sense that there are no label errors per se. However, the ground truth is unrealistically detailed in comparison to annotation created by a human. Also DNNs are not able to produce such fine grained annotations. Hence we smooth the labels such that objects closer to the ego vehicle diffuse over objects farther away. For more details on this see \cref{App:A}. 

The Cityscapes dataset is a high quality dataset that at least upon visual inspection contains a rather small number of label errors. We report numbers on this in \cref{sec:numexp}. Hence we also create a benchmark by synthetically inducing label errors by dropping polygons from the ground truth, for further details see \cref{App:A}.
%in this dataset. , while the benchmark will only exhibit a moderate bias due to real label errors. Still, the application to real data justifies this procedure in our opinion. 

\paragraph{Evaluation protocol.}
In order to understand the connection of network performance and label error detection performance, we state the underlying network's performance in terms of mean intersection over union ($\mathrm{mIoU}$). The $\mathrm{mIoU}$ is the mean over all classes of the intersection over union ($\mathrm{IoU}$) where the latter for each class is computed as $\mathit{tp}/(\mathit{tp}+\mathit{fp}+\mathit{fn})$ where $\mathit{tp}$ denotes the number of true positive pixels, $\mathit{fp}$ the number of false positive pixels and $\mathit{fn}$ the number of false negative pixels of the given class within a given test set.

For benchmark purposes, assume the existence of a clean ground truth without label errors. For a given image, the set of clean ground truth connected components is denoted by $K_c$. Let $K_\ell$ % := K_c \setminus K $ 
denote the label error ground truth that registers all label errors in terms of connected components. Given another set of connected components $K'$ that constitutes label error proposals, let $m' : K' \to [0,1]$ be some probabilistic label error detection method. For a given decision threshold $t$ let $\hat{K}_\ell(t) := \{ \hat{k} \in K' : m'(\hat{k}) \geq t \}$ denote the set of all connected components predicted by $m'$.

We repeat the construction from \cref{sec:method}
replacing $\hat{K}$ by $\hat{K}_\ell(t)$ and $K$ by $K_\ell$ and proceed analogously to \cref{sec:method} with the same threshold $\tau$ to define $\mathrm{TP}(t)$, $\mathrm{FP}(t)$ and $\mathrm{FN}(t)$ in the context of label error detection. That is, a $\mathrm{TP}$ is a correctly detected label error, an $\mathrm{FP}$ is an incorrect label error detection (a false discovery) and an $\mathrm{FN}$ is an overlooked label error.
As additional evaluation metrics, we consider precision $\mathrm{prec}(t) = \frac{\mathrm{TP}(t)}{\mathrm{TP}(t)+\mathrm{FP}(t)}$, recall $\mathrm{rec}(t) = \frac{\mathrm{TP}(t)}{\mathrm{TP}(t)+\mathrm{FN}(t)}$, the $F_1$-score $F_1(t) = \frac{2\mathrm{TP}(t)}{2\mathrm{TP}(t)+\mathrm{FP}(t)+\mathrm{FN}(t)}$ and for $\mathrm{PRC}(\mathrm{rec}(t)) = \mathrm{prec}(t) $ we define the average precision $\mathrm{AP} = \int_{0}^1 \mathrm{PRC}(r) \, \mathrm{d}r $. Whenever the threshold $t$ is not of interest or pre-specified, we will omit the argument $t$ in the above definitions.

Technically, the evaluation code we provide allows for label error detection confidences of $m$ to be provided on pixel level as well. In that case, also the threshold $t$ is applied to the estimated probabilities (label error detection scores) on pixel level.

\section{Numerical Experiments} \label{sec:numexp}

\paragraph{Experiment setup.}

For numerical experiments we use a Deeplabv3+ \cite{Chen2018EncoderDecoderWA} architecture with a WideResNet38 \cite{Wu2019WiderOD} backbone, and Nvidia's multi-scale attention approach \cite{Tao2020HierarchicalMA} in combination with an HRNet-OCR trunk \cite{Sun2019HighResolutionRF} to make use of the current state-of-the-art networks in semantic segmentation.
Both networks are trained on our Cityscapes and CARLA datasets on different perturbation levels corresponding to label perturbation probability $\hat{p}=0$, $0.1$, $0.25$ and $0.5$.
For evaluation we consider two scenarios.

\emph{Experiment 1)} we assume to have a chunk of very well labeled data for training, i.e., the training data has $\hat{p}=0$. Furthermore we assume that the rest has not been reviewed extensively, i.e., we test the label error performance for $\hat{p}=0.5$. Note that one choice of $\hat{p}$ for evaluation in this scenario is sufficient. Due to the independence of the Bernoulli trials, we can expect that for a fixed network $f$ and a fixed threshold $t$, varying $\hat{p}$ will scale $\mathrm{TP}$ and $\mathrm{FN}$ proportionally to each other  while $\mathrm{FP}$ remains constant (i.e., recall remains constant while precision increases accordingly).

\emph{Experiment 2)} we assume that all data labeled data is of the same quality, i.e., we train the network on the original data for a given value of $\hat{p}$ and evaluate w.r.t.\ label error detection for the same value of $\hat{p}$. Also here the independence of Bernoulli trials makes further cross evaluations for choices of $\hat{p}$ obsolete.

In case of Cityscapes, we train on the pre-defined official training set and state results for the official validation set. For CARLA we choose town $1$--$4$ as training maps and town $5$ for validation to avoid as much correlation in our split as possible.

In addition to the evaluation on synthetically generated label errors in the upcoming \cref{sec:syntherrors}, we provide findings on real label errors in frequently used semantic segmentation datasets along with sample statistics on $\mathrm{TP}$ and $\mathrm{FP}$ in \cref{sec:realerrors}.

\subsection{Experiments with Induced Label Errors} \label{sec:syntherrors}
% In training we make use of Nvidia's Pytorch implementation of the Deeplab and their hierarchical multi-scale attention architecture \cite{NvidiaFramwork}.
% For both networks, we employ $8$ GPUs for training with a batch size of $2$ per GPU, momentum $0.9$ and weight decay $10^{-4}$. We use a polynomial learning rate \cite{liu2015parsenet} 
% with an exponent of $2.0$, an
% initial learning rate of $0.005$ for the Deeplab network and of $0.01$ for the multi-scale attention net, and train for $200$ epochs. For the attention net we scale the input image by a factor of $0.5$, $1$, and $2$.
% In addition, we use class uniform sampling in the data loader to improve the results in case of an nonuniform data distribution.
% For the prediction of the label masks, we use the final checkpoint of the training process to reduce the amount of adaption to label errors in the validation set.

For the evaluation, we have fixed $\tau=0.25$ as $\mathrm{sIoU}$ threshold for determining $\mathrm{TP}_o$, $\mathrm{FP}_o$ and $\mathrm{FN}_o$. On the other hand, we consider 
only predicted components that have no intersection of same class with the ground truth annotation mask.
%\MR{We introduce a confidence threshold $t$ on the meta classifiers output $m(\hat{k})$, $\hat{k} \in \hat{K}$ to control the sensitivity of our label error detector.} 
We report numbers for $\mathrm{TP}$, $\mathrm{FP}$ and $\mathrm{FN}$, precision ($\mathrm{prec}$) and recall ($\mathrm{rec}$) for a choice of $t$ that maximizes the $F_1$ score. In addition we compare our 
approach against two naive baseline methods. Assuming that we are able to detect every label error by comparing the perturbed ground truth label mask and the input image, in baseline method $1$ we review every
connected component larger than $250$ pixel of the perturbed label mask to evaluate if a component is missing.
In baseline method $2$ we review every false positive component $\mathrm{FP}_o$ produced by the DNN without any further probabilistic considerations.
%In addition to the standard metrics, we also discuss the highest possible recall if we 
%allow the precision to decrease to $10\%$. This value shall be denoted with Rec$10$ and represents the highest possible amount of found label errors.

\paragraph*{Results on CARLA.}
Our CARLA dataset consists of $6000$ images of size $1024\times 2048$ pixels with $17$ different classes. The first $4800$ images we recorded in $4$ different towns provided by CARLA and are used for training. The fifth town with $1200$ image is used for the evaluation. We train the attention net with the original image size while using crops of size $800\times 800$ pixels for training the Deeplab network. We use approximately half of the perturbed validation set to train the meta classifier $m$ to estimate $P(\mathrm{TP}_o)$ and search in the other half for label errors. 
This split remains the same across all perturbation levels. For this dataset we have $154$, $514$, and 
$1151$ label errors for $\hat{p}=0.1$, $0.25$, and $0.5$, respectively. The calculation of the $\mathrm{mIoU}$ is always done w.r.t.\ an unperturbed version of the validation set.

First we consider the results of experiment $1$ in which we train the networks on an unperturbed / clean training set and aim to find errors in perturbed validation sets with different perturbation rates. With both networks we find most of the label errors we induced in the validation set; see \cref{table:carla2}. 
%In particular for the Nvidia multiscale attention architecture these results are remarkable.
With the Nvidia multiscale attention architecture, we find at least $76.28\%$ of all label errors for all rates $\hat{p}$ while having a precision of over $72\%$. The results for the Deeplab net are lower compared to the Nvidia net. However, we also find most of the label errors with approximately the same precision. Compared to the baseline method we see that our method provides a high improvement in the precision while the decrease in recall 
is small. This observation is particularly pronounced for the lowest perturbation rate of $\hat{p}=0.1$ where we for the attention net gain additional $49.33$ percent points (pp) in the precision and only lose $3.87$ pp in the recall.
%\MR{\emph{MR: please check the previous sentence for correctness as I changed it a lot}}. 
For a higher rate
the difference between our method and the baseline is rather small using Nvidias's architecture. We attribute this to the synthetic nature of this dataset and the fact that the networks are trained on clean training sets.

The overall detection capabilities of our method are rated by the $\mathrm{AP}$ scores obtained by varying the values of $t$. For the attention net, these $\mathrm{AP}$ scores are stable for all rates $\hat{p}$ and even slightly increase for a rate $\hat{p}=0.5$. For the Deeplab net we can observe the same improvement for larger rates $\hat{p}$. Most importantly, our method scales very well 
with the number of label errors in the dataset, being functional in presence of a few label errors as well as when facing many label errors. %While the former can be attributed to be performance of our method and that of the DNN, the latter is also a consequence of the robustness of DNNs w.r.t.\ label errors.}

%Overall, in this scenario our method provides strong results which seems not to suffer from higher perturbation levels.

\begin{table}[h]
\resizebox{0.47\textwidth}{!}{
\label{table:carla2}
\begin{tabular}{@{}l@{\hskip 0.3cm}rrrrrrrr@{}}
\toprule
    & $\mathrm{mIoU}$ & $\mathrm{TP}$   & $\mathrm{FN}$   & $\mathrm{FP}$ & $\mathrm{AP}$ & $\mathrm{Prec}$ & $\mathrm{Rec}$ & $\mathrm{F1}$ \\ \hline\hline
    \hspace*{0.25cm}\footnotesize{\emph{$\hat{p}=0.1$}} \\
    Nvidia Attn. Net & 78.61 & 131 & 24 & 12 & 66.10 & 91.61 & 84.52 & 87.92\\
    DeepLabV3+ 38 & 71.52 & 93 & 62 & 39 & 44.49 & 70.46 & 60.00 & 64.81\\
    Base. 1 & -- & 155 & 0 & 39667  & -- & 00.39 & 100.00 & 00.78\\
    Base. 2: Nvidia & 78.61 & 137 & 18 & 187 & -- & 42.28 & 88.39 & 57.20 \\
    Base. 2: Deeplab & 71.52 & 108 & 47 & 794 & -- & 11.97 & 69.68 & 20.44 \\ 
    ~ \\
    \hspace*{0.25cm}\footnotesize{\emph{$\hat{p}=0.25$}} \\
    Nvidia Attn. Net & 78.61 & 401 & 113 & 154 & 67.87 & 72.25 & 78.02 & 75.02 \\
    DeepLabV3+ 38 & 71.52 & 278 & 236 & 103 & 54.87 & 72.97 & 54.09 & 62.12 \\
    Base. 1 & -- & 514 & 0 & 39055  & -- & 01.30 & 100.00 & 02.56\\
    Base. 2: Nvidia & 78.61 & 406 & 108 & 321 & -- & 55.85 & 78.99 & 65.43 \\
    Base. 2: Deeplab & 71.52 & 319 & 195 & 879 & -- & 26.63 & 62.06 & 37.27 \\ 
    ~ \\
    \hspace*{0.25cm}\footnotesize{\emph{$\hat{p}=0.5$}} \\
    Nvidia Attn. Net & 78.61 & 878 & 273 & 284 & 71.70 & 75.56 & 76.28 & 75.92 \\
    DeepLabV3+ 38 & 71.52 & 654 & 497 & 213 & 60.54 & 75.43 & 56.82 & 64.82 \\
    Base. 1 & -- & 1151 & 0 & 37951 & -- & 02.94 & 100.00 & 05.71\\
    Base. 2: Nvidia & 78.61 & 885 & 266 & 438 & -- & 66.89 & 76.89 & 71.54 \\
    Base. 2: Deeplab & 71.52 & 715 & 436 & 993 & -- & 41.86 & 62.12 & 50.02 \\ 
\bottomrule
\end{tabular}}
\vspace{0.5em}
\caption{Results for label error detection trained on clean and validated on perturbed CARLA datasets.}
\end{table}

For experiment $2$ we trained the networks on perturbed training sets and validated our method also on perturbed validation sets with the same perturbation rate $\hat{p}$. The obtained results are of similar quality as for experiment 1.
%Again, we obtain very strong results for both nets.
With the attention net, for a rate of $\hat{p}=0.1$ we find $87.1\%$ of all label errors and $75.49\%$ for a rate of $0.25$ while having a very high precision of $91.22\%$ and $71.19\%$. At a rate of $\hat{p}=0.5$ our method still 
achieves descent results where, however, the recall decreases significantly compared to the lower rates. This signals that we are not able to find all label errors at once at higher rates $\hat{p}$, but also indicates that an iterative procedure of DNN training and label error cleaning might be an interesting direction.

As for experiment $1$ we observe that our method provides a significantly higher precision while the decrease in the recall is minor.

%Similarly to Scenario $1$ the AP values are improving for higher values which indicates again that our methods scales well with increasing perturbation rates.

\begin{table}[h]
\resizebox{0.47\textwidth}{!}{
\label{table:carl1}
\begin{tabular}{@{}l@{\hskip 0.3cm}rrrrrrrrr@{}}
\toprule
    & $\mathrm{mIoU}$ & $\mathrm{TP}$   & $\mathrm{FN}$   & $\mathrm{FP}$ & $\mathrm{AP}$ & $\mathrm{Prec}$ & $\mathrm{Rec}$ & $\mathrm{F1}$ \\ \hline\hline
    \hspace*{0.25cm}\footnotesize{\emph{$\hat{p}=0.1$}} \\
    Nvidia Attn. Net & 79.16 & 135 & 20 & 13 & 62.57 & 91.22 & 87.10 & 89.11 \\
    DeepLabV3+ 38 & 70.58 & 82 & 73 & 66 & 42.35 & 55.41 & 52.90 & 54.13 \\
    Base. 1 & -- & 155 & 0 & 39667  & -- & 00.39 & 100.00 & 00.78\\
    Base. 2: Nvidia & 79.16 & 138 & 17 & 148 & -- & 48.25 & 89.03 & 62.59 \\
    Base. 2: Deeplab & 70.58 & 103 & 52 & 987 & -- & 09.45 & 66.45 & 16.55 \\ 
    ~ \\
    \hspace*{0.25cm}\footnotesize{\emph{$\hat{p}=0.25$}} \\
    Nvidia Attn. Net & 79.54 & 388 & 126 & 157 & 66.85 & 71.19 & 75.49 & 73.28 \\
    DeepLabV3+ 38 & 70.44 & 251 & 263 & 112 & 53.94 & 69.15 & 48.83 & 57.24 \\
    Base. 1 & -- & 514 & 0 & 39055   & -- & 01.39 & 100.00 & 02.56\\
    Base. 2: Nvidia & 79.54 & 395 & 119 & 225 & -- & 63.71 & 76.85 & 69.66 \\
    Base. 2: Deeplab & 70.44 & 272 & 242 & 709 & -- & 27.73 & 52.92 & 36.39 \\
    ~ \\
    \hspace*{0.25cm}\footnotesize{\emph{$\hat{p}=0.5$}} \\
    Nvidia Attn. Net & 77.42 & 631 & 520 & 378 & 62.28 & 62.54 & 54.82 & 58.43 \\
    DeepLabV3+ 38 & 70.25 & 583 & 568 & 334 & 55.74 & 63.58 & 50.65 & 56.38 \\
    Base. 1 & -- & 1151 & 0 & 37951 & -- & 02.94 & 100.00 & 05.71\\
    Base. 2: Nvidia & 77.42 & 632 & 519 & 431 & -- & 59.45 & 54.91  & 57.09 \\
    Base. 2: Deeplab & 70.25 & 643 & 508 & 801 & -- & 44.53 & 55.86 & 49.56 \\
\bottomrule
\end{tabular}}
\vspace{0.5em}
\caption{Results for label error detection trained and validated on perturbed CARLA datasets.}
\end{table}

\paragraph*{Results on Cityscapes.}
Cityscapes contains $5000$ high resolution images annotated into $19$ classes. For the evaluation, we train on the pre-defined training set which consists of $2975$ images with a resolution of 
$1024 \times 2048$ pixels and evaluate on the pre-defined validation set containing $500$ images of the same resolution. In analogy to our CARLA experiments, for the attention net we use the original image size, i.e., we do not use random cropping. For Deeplab, we use a crop size of $800\times 800$ pixels. Again, we use one half of the validation set to train the meta classifier $m$ and aim at finding label errors in the other half, where this split is identical for all perturbation rates $\hat{p}$. The latter are chosen in analogy to the CARLA experiments.
%In both datasets we induced label errors with a probability of $\hat{p} = 0.1,\ 0.25$, and $0.5$ (see section \ref{sec:datasets}). 
In total we have $166$, $381$, and $746$ label errors induced in Cityscapes for $\hat{p} = 0.1,\ 0.25$, and $0.5$, respectively.

\Cref{table:cs2} contains the results for experiment $1$ in which we train on clean data. In addition to the models trained by ourselves, we also use current state-of-the-art (sota) weights provided by \cite{DeeplabWeights} and \cite{NvidiaFramwork}. %Again, for both networks we get remarkable results. 
For perturbation rates of $\hat{p}=0.1$ and $\hat{p}=0.25$ we find $43.57\%-57.83\%$ of all label errors with the attention net (sota and self trained) while we only have to look at $2$ to $3$ 
candidates on average to find one label error. In addition, the recall increases significantly to $60.72\%$ and $65.28\%$ for a perturbation rate of $\hat{p}=0.5$ for the Nvidia architecture with self trained and sota weights. The precision of our approach also increases for higher perturbation rates and varies between $40.41\%$ and $64.41\%$ for the Deeplab net and between $36.09\%$ and $62.50\%$ for Nvidia's multiscale attention net.

We note that for the Cityscapes dataset, the false positive label errors (i.e., the label error predictions being identified as $\mathrm{FP}$ according to our benchmark) also contain real label errors which were not induced by us and are therefore actually to be counted as true positives. We study this finding more precisely in \cref{sec:realerrors}. Noteworthily, as the $\mathrm{FP}$ count remains under control, this signals that the number of label errors in Cityscapes can be expected to be rather moderate. 

For combination of dataset and experiment, the baseline method 2 (and obviously method 1 as well) provides overall very low precision scores. In particular for $\hat{p}=0.1$ we only obtain precision scores between $2.90\%$ and $8.18\%$. 
The loss in recall is larger compared to 
the CARLA dataset. The fact that our method achieves a high recall at way higher precision than the baselines demonstrates the significantly higher efficiency of our method.
%we still obtain high recall with our method it appears to be significantly more efficient to apply our method once or multiple times instead of reviewing every false positive segment predicted by the DNN.
%It is also worth noting that the architectures we chose are very robust to label errors in training sets as the $\text{mIoU}$ of all models are stable across all perturbation rates. \MR{This finding is in accordance to the literature \cite{}.}

%Since we have somewhat fixed the amount of true positive, the precision improves 
%significantly while the recall decreases for increasing $\hat{p}$.
%with the same amount of false positives we found more label errors and therefore obtained improved precision and recall values for each model.
\begin{table}[h]
\resizebox{0.47\textwidth}{!}{
\begin{tabular}{@{}l@{\hskip 0.3cm}rrrrrrrrr@{}}
\toprule
    & $\mathrm{mIoU}$ & $\mathrm{TP}$   & $\mathrm{FN}$   & $\mathrm{FP}$ & $\mathrm{AP}$ & $\mathrm{Prec}$ & $\mathrm{Rec}$ & $\mathrm{F1}$\\ \hline\hline
    \hspace*{0.25cm}\footnotesize{\emph{$\hat{p}=0.1$}} \\
    Nvidia Attn. Net & 83.32 & 96 & 70 & 170 & 35.19 & 36.09 & 57.83 & 44.44 \\
    DeepLabV3+  & 79.37 & 59 & 107 & 87 & 23.89 & 40.41 & 35.54 & 37.82 \\
    Nvidia Attn. Net sota & 86.82 & 72 & 94 & 91 & 33.94 & 44.17 & 43.37 & 43.77 \\
    DeepLabV3+ sota & 83.50 & 86 & 90 & 61 & 36.38 & 58.50 & 51.81 & 54.95 \\
    Base. 1 & -- & 166 & 0 & 18458 & -- & 00.89 & 100.00 & 01.76\\
    Base. 2: Nvidia & 83.32 & 127 & 39 & 2117 & -- & 05.66 & 76.51 & 10.54 \\
    Base. 2: Deeplab & 79.37 & 110 & 56 & 3684 & -- & 02.90 & 66.27 & 05.56 \\ 
    Base. 2: Nvidia sota & 86.82 & 126 & 40 & 1414 & -- & 08.18 & 75.90 & 14.77 \\
    Base. 2: Deeplab sota & 81.40 & 128 & 38 & 1805 & -- & 06.62 & 77.11 & 12.20 \\ 
    ~ \\
    \hspace*{0.25cm}\footnotesize{\emph{$\hat{p}=0.25$}} \\
    Nvidia Attn. Net & 83.32 & 185 & 196 & 111 & 51.82 & 62.50 & 48.56 & 54.65 \\
    DeepLabV3+ & 79.37 & 152 & 229 & 95 & 49.94 & 61.54 & 39.90 & 48.41 \\
    Nvidia Attn. Net sota & 86.82 & 194 & 187 & 141 & 51.40 & 57.91 & 50.92 & 54.19 \\
    DeepLabV3+ sota & 81.40 & 228 & 153 & 126 & 49.82 & 64.41 & 59.84 & 62.04 \\
    Base. 1 & -- & 381 & 0 & 18037 & -- & 02.07 & 100.00 & 04.06\\
    Base. 2: Nvidia & 83.32 & 271 & 110 & 2140 & -- & 11.24 & 71.13 & 19.41 \\
    Base. 2: Deeplab & 79.37 & 239 & 142 & 3742 & -- & 06.00 & 62.73 & 10.96 \\ 
    Base. 2: Nvidia sota & 86.82 & 288 & 93 & 1388 & -- & 17.18 & 75.59 & 28.00 \\
    Base. 2: Deeplab sota & 83.50 & 292 & 89 & 1840 & -- & 13.70 & 76.65 & 23.24 \\ 
    ~ \\
    \hspace*{0.25cm}\footnotesize{\emph{$\hat{p}=0.5$}} \\
    Nvidia Attn. Net & 83.32 & 453 & 293 & 327 & 53.81 & 58.08 & 60.72 & 59.37 \\
    DeepLabV3+ & 79.37 & 343 & 403 & 246 & 47.69 & 58.23 & 45.98 & 51.39 \\
    Nvidia Attn. Net sota      & 86.82 & 487 & 259 & 371 & 56.26 & 56.76 & 65.28 & 60.72 \\
    DeepLabV3+ sota & 81.40 & 481 & 265 & 325 & 61.42 & 59.68 & 64.48 & 61.98 \\
    Base. 1 & -- & 746 & 0 & 17367 & -- & 04.12 & 100.00 & 07.91\\
    Base. 2: Nvidia & 83.32 & 540 & 206 & 2106 & -- & 20.41 & 72.39 & 31.84 \\
    Base. 2: Deeplab & 79.37 & 473 & 273 & 3799 & -- & 11.07 & 63.40 & 18.85 \\ 
    Base. 2: Nvidia sota & 86.82 & 567 & 179 & 1448 & -- & 28.14 & 76.01 & 41.07 \\
    Base. 2: Deeplab sota & 83.50 & 563 & 183 & 1860 & -- & 23.24 & 75.47 & 35.53 \\ 
\bottomrule
\end{tabular}}
\vspace{0.5em}
\caption{Results for label error detection trained on clean and validated on perturbed Cityscapes datasets.}
\label{table:cs2}
\end{table}

\Cref{table:cs1} summarizes our results on Cityscapes for experiment $2$ where we trained the network and the label error detection on perturbed datasets. We still find most of the label errors we induced. However, as one would assume, training on unperturbed data causes a slight deterioration to our method.
With the attention net and perturbation rates of $\hat{p}=0.1$ as well as $\hat{p}=0.25$, we find approximately between $54\%$ and $56\%$ of all label errors while we only have to review on average $2$ to $3$ candidates per label error. For a perturbation rate of 
$\hat{p}=0.5$, that ratio decreases to roughly $2$ candidate reviews per label error. On the other hand, we can again observe a drop in recall to $45.58\%$. Furthermore, as already observed in our first two experiments with CALRA data, the $\mathrm{AP}$ scores are improving for larger perturbation rates for both networks.

\begin{table}[h]
\label{table:cs1}
\resizebox{0.47\textwidth}{!}{
\begin{tabular}{@{}l@{\hskip 0.3cm}rrrrrrrrr@{}}
\toprule
    & $\mathrm{mIoU}$ & $\mathrm{TP}$   & $\mathrm{FN}$   & $\mathrm{FP}$ & $\mathrm{AP}$ & $\mathrm{Prec}$ & $\mathrm{Rec}$ & $\mathrm{F1}$ \\ \hline\hline
    \hspace*{0.25cm}\footnotesize{\emph{$\hat{p}=0.1$}} \\
    Nvidia Attn. Net & 82.71 & 93 & 73 & 153 & 36.36 & 37.80 & 56.02 & 45.15 \\
    DeepLabV3+   & 79.06 & 43 & 123 & 46 & 20.50 & 48.31 & 25.90 & 33.73 \\
    Base. 1 & -- & 166 & 0 & 18458 & -- & 00.89 & 100.00 & 01.76\\
    Base. 2: Nvidia & 82.71 & 124 & 42 & 1981 & -- & 05.89 & 74.70 & 10.92 \\
    Base. 2: Deeplab & 79.06 & 103 & 63 & 3510 & -- & 02.85 & 62.05 & 05.45 \\ 
    ~ \\
    \hspace*{0.25cm}\footnotesize{\emph{$\hat{p}=0.25$}} \\
    Nvidia Attn. Net & 82.74 & 208 & 173 & 216 & 46.34 & 49.06 & 54.59 & 51.68 \\
    DeepLabV3+   & 78.84 & 146 & 235 & 198 & 29.90 & 42.44 & 38.32 & 40.28 \\
    Base. 1 & -- & 381 & 0 & 18037 & -- & 02.07 & 100.00 & 04.06\\
    Base. 2: Nvidia & 82.74 & 265 & 116 & 1823 & -- & 12.69 & 69.55 & 21.47 \\
    Base. 2: Deeplab & 78.84 & 232 & 149 & 5335 & -- & 04.17 & 60.89 & 07.80 \\
    ~ \\
    \hspace*{0.25cm}\footnotesize{\emph{$\hat{p}=0.5$}} \\
    Nvidia Attn. Net & 82.68 & 340 & 406 & 288 & 48.52 & 54.14 & 45.58 & 49.49 \\
    DeepLabV3+   & 78.67 & 290 & 456 & 422 & 34.02 & 40.73 & 38.87 & 39.78 \\
    Base. 1 & -- & 746 & 0 & 17367 & -- & 04.12 & 100.00 & 07.91\\
    Base. 2: Nvidia & 82.68 & 465 & 281 & 1747 & -- & 21.02 & 62.33 & 31.44 \\
    Base. 2: Deeplab & 78.67 & 436 & 310 & 4078 & -- & 09.66 & 58.45 & 16.58 \\
\bottomrule
\end{tabular}}
\vspace{0.5em}
\caption{Results for label error detection trained and validated on perturbed Cityscapes datasets.}
\end{table}

\subsection{Label Errors in Frequently used Datasets}\label{sec:realerrors}
In this section, we use our method to find real label errors in popular semantic segmentation datasets. We employ Nvidia's multi-scale attention net with an HRNet-OCR trunk. For each dataset, we either use pretrained state-of-the-art weights or train the network on the predefined training sets ourselves. 
% For training, we employ the same hardware and set the hyperparameters in the same way as described in \ref{sec:syntherrors} where only the amount of epochs is adjusted to the size of the respective dataset. 
We examine the validation set of each dataset (except for Cityscapes where we also consider the training set), where we use one half of the set to train the label error detection and search in the other half for real label errors.
Then we switch the roles of the splits, such that we search in both split for label errors. For each split we validate the $100$ $\mathrm{FP}_o$-components $\hat{k}$ (having no intersection with the ground truth) with the highest estimated $m(\hat{k}) \approx P(\mathrm{TP}_o)$ (which is achieved by sorting $m(\hat{k})$). Hereby, we strictly follow the label policies of the datasets and only confirm predictions as real label errors when there is no doubt about their correctness, i.e., we proceed rather conservatively.

\paragraph*{Results on Cityscapes.}
For Cityscapes, we examine the training set and the validation set, individually. In addition, we study the classes 
person, rider/bicycle and vehicles independently of the classes of traffic lights and road signs in order to avoid that the classifier gives us an unbalanced amount traffic lights and road signs predictions to validate.
We consider $75$ candidates of the first set of classes and $25$ of latter classes for each split. For Cityscapes the sota weights provide an $\mathrm{mIoU}$ of $86.82\%$. In the training set we achieve a precision of $57.5\%$ which amounts to $115$ true label errors and $85$ false discoveries within the $200$ most probable label error candidates with highest meta classification score within the $2975$ images of the training set; see \cref{table:error_study}. 
%We note again that we only study the $200$ predictions which are most likely to be label errors according to the meta classifier $m$.

% For the classes of person, rider/bicycle and vehicles we found $90$ true positives, $60$ false positive, and a precision of $60\%$. For traffic lights and road signs we achieve a precision of $50\%$ and found $25$ label errors in $50$ predictions.

\begin{figure}[h!]
    \center
    {\includegraphics[width=.47\textwidth]{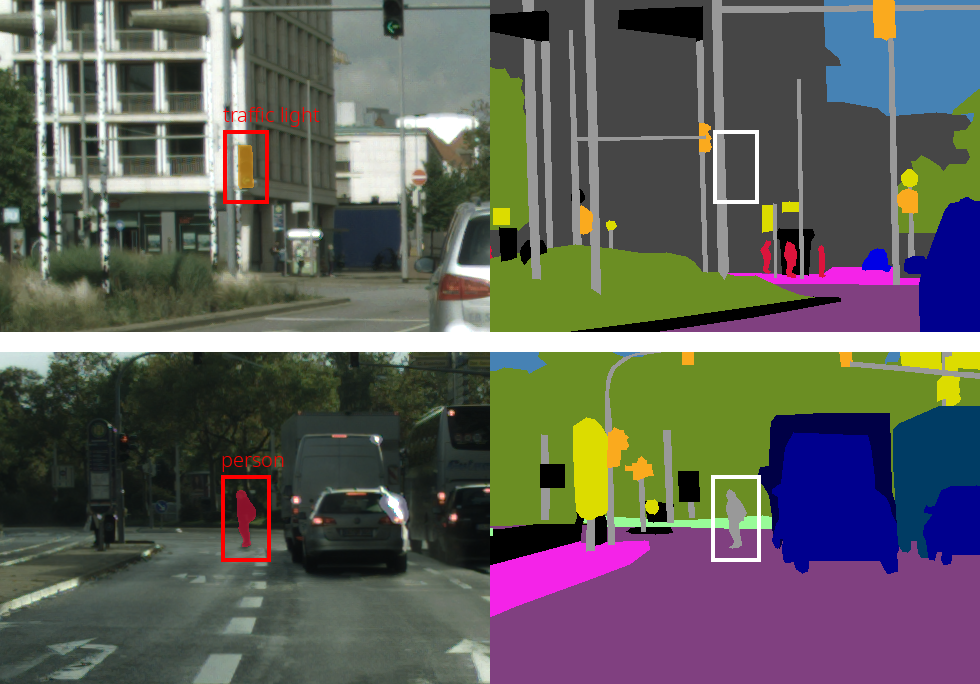}}
    \vspace{.5em}
    \caption{Two examples of label errors detected by our method, one example per row. Left: prediction of our label error detection method; right: ``ground truth'' annotation. Our method is able to find both, overlooked and flipped labels.}
    \label{fig:cs}
\end{figure}

On the validation set our method achieves a precision of $53\%$ where we found $106$ label errors when considering $200$ candidates.
% For the classes of person, rider/bicycle and vehicles we have a precision of $55\%$ and of $46\%$ for the classes of  traffic lights and road signs, respectively. 
The results for the validation set are a slightly inferior compared to the results 
for the training set which is to be expected as the validation set only contains $500$ images while we examine the same amount of discoveries from our method. 
For a class-wise breakdown of these results, see  \cref{table:cs_study_val} in \cref{app:label_breakdown}. Two examples of identified label errors are presented in \cref{fig:cs}.

\begin{table}[h]
    \begin{center}
    \label{table:error_study}
    \resizebox{0.4\textwidth}{!}{
    \begin{tabular}{ lrrrr }
    \toprule
     Dataset & $\mathrm{mIoU}$ & $\mathrm{TP}$ & $\mathrm{FP}$ & $\mathrm{Prec}$ \\
    \hline\hline
    Cityscapes Training & 86.82 & 115 & 85 & 57.50 \\
    Cityscapes Validation & 86.82 & 106 & 94 & 53.00 \\ 
    PascalVOC Validaiton & 78.03 & 95 & 105 & 47.50 \\
    Coco Validation & 28.20 & 134 & 66 & 67.00 \\
    ADE20K & 43.12 & 110 & 90 & 55.00 \\ 
    \bottomrule
    \end{tabular}
    }
    \vspace{0.5em}
    \caption{Precision of our approach for different datasets.}
    \end{center}
\end{table}

Summarizing the Cityscapes results, by reviewing only 
$400$ candidates conservatively, we already identified $231$ label errors. The precision of $53\%$ for the small validation set indicates that there may remain further label errors which can be approached further by using our method.

\paragraph*{Results on PascalVOC.}
The PASCAL Visual Object Classes (VOC) 2012 dataset \cite{Everingham15} contains $20$ object categories of a wide variety. The training set consists of $8497$ and the validation set of $2857$ images. 
The attention net 
% is trained for $200$ epochs and 
achieves a $\mathrm{mIoU}$ of $78.03\%$ on the validation set. 
For the current and all subsequent datasets, we only study the validation set as we expect a vast amount of label errors therein. Conversely to Cityscapes, we do not additionally filter or split the classes. 
For the evaluation, we consider every class and again examine the $200$ candidates with highest meta classification probability.

\begin{figure}[h!]
    \center
    {\includegraphics[width=.47\textwidth]{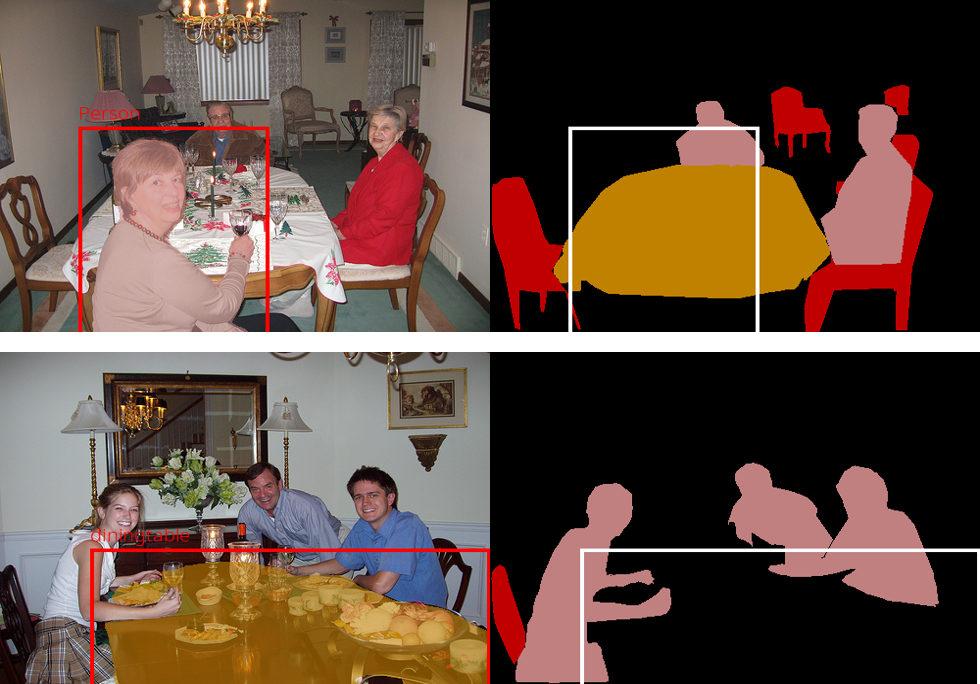}}
    \vspace{.5em}
    \caption{Two detected label errors in PascalVOC. The visualization scheme follows the one of \cref{fig:cs}.}
    \label{fig:ps}
\end{figure}

We identified $95$ label errors and obtained a precision of $47.5\%$; see \cref{table:error_study}. One might expect that a high $\mathrm{IoU}$ for a specific class would result in a high precision for this class. 
However, our results contradicts this intuition. While this is true for the class of Person for which we achieve a precision of $84\%$, the opposite is true for almost every other class.
For a detailed discussion on these results we refer to \cref{app:label_breakdown}. Two exemplary label errors are shown in \cref{fig:ps}
% The best results are achieved for the class Person with a precision of $84\%$. 
%
% However, our results contradicts this assumption. While this is true for the class of Person the opposite is true for almost every other class. For example, according 
% to the IoU values, the attention net should be able to find cats and dogs quite decently. In our tests the net often confuses these classes, while the meta classifier agrees with the prediction of 
% the net. This results in a lot of false positives. Merely $1$ out of $11$ cat label error predictions were correct. At the same time, our method is still able to find label errors in classes with low IoUs. For example, we
% only have a IoU of $51.97$ for the chair class but still could find $17$ label errors with a precision of $61\%$. We can observe the same for the class Dining table in which we found $10$ label errors with 
% a precision of $0.63$ and an IoU of $51.47$.
% This indicates that the low mIoU might be partially caused by a poor labeling quality of the dataset.
% \MR{MR: here I would like to discuss and get a deeper understanding. What role does the taxonomy and overconfidence maybe play?}

\paragraph*{Results on COCO-Stuff.}
The COCO-Stuff dataset \cite{cocostuff} is the largest dataset we examine, containing $118$K training images and $5$K validation images of variable size and $93$ total categories. 
% Due to the size of the training set, we train for $100$ epochs with 
% a crop size of $600\times 600$ pixels. 
On the validation set, the net achieves an $\mathrm{mIoU}$ of $28.20\%$ and our approach obtained a high precision of $67\%$. This amounts to $134$ true positives within $200$ candidates from $36$ class. For this dataset we also observed that a low $\mathrm{IoU}$ does not automatically imply a low precision. Again we refer to
\cref{app:label_breakdown} for more details. Note that for this dataset we were not able find a documentation that specifies the annotation policy precisely. Therefore, in our evaluation, we used the class descriptions.

% As some classes are not present in the validation set, we obtained a few NaN values for the 
% corresponding IoU; see the full Table \textcolor{red}{appendix}. In addition, some classes appear to be underrepresented in the dataset, therefore it is not clear how meaningful these IoU values are. 

% The best results are given for the classes of 
% Building-other, Grass, Playingfield, Sky-other and Tree with precision values significantly above $50\%$ and a representative amount of predictions. ; see e.g.\ the results for class Building-other. 

%\begin{figure}[h!]
%    \center
%    \caption{Two label error examples in COCO.}
%    {\includegraphics[width=.47\textwidth]{../images/coco_fig.png}}
%    \label{fig:coco_fig}
%\end{figure}

\paragraph*{Results on ADE20K.}
ADE20K \cite{ade1, ade2} consists of $27$K images annotated in $151$ classes. We trained for $200$ epochs and a crop size of $600\times 600$ pixels on $25$K training images and applied our method to the validation set of 
$2$K images, where we obtain a precision $52.5\%$ with an $\mathrm{mIoU}$ of $43.12\%$. \Cref{table:error_study} shows the results for this dataset. For the further details see \cref{app:label_breakdown}.
We found $104$ label errors in $200$ predictions across $58$ classes. We note that the class definitions of this dataset are in part not sufficiently distinct. In addition, since we were not able to find class descriptions for this dataset, we had to infer from the ground truth annotation which objects the classes represent. 
This led to a significant amount of false positives since we were regularly not able to evaluate confidently whether the prediction is a true positive or not. 

% This applies in particular to 
% the class ``building''. In ADE20K exists also the class house which seems to describe buildings for human habitation. For the net it was difficult to differentiate between these two classes, hence it predicts most 
% of the buildings/houses as buildings. This leads to high IoU values for the class building, but a low one of $25.05$ for the class house and a precision of $0\%$ for the building class. However, despite the mentioned issues, the results for ADE20K are comparable to those obtained for the other datasets.

%\begin{figure}[h!]
%    \center
%    \caption{Two label error examples in ADE20K.}
%    {\includegraphics[width=.47\textwidth]{../images/ade_fig.png}}
%    \label{fig:ade_fig}
%\end{figure}

\section{Conclusion \& Outlook}

In this work, we demonstrated how uncertainty quantification leverages the detection of label errors in semantic segmentation datasets via deep learning. We find good trade-offs of human labor and discovery rates of label errors, therefore enabling the efficient quality improvement for semantic segmentation datasets. In the future, we plan to develop measures and estimators for overall dataset quality. Furthermore, we plan to extend our method to detect falsely induced connected components (e.g.\ a piece of sky within the street) that are likely to be overlooked by the DNN, thus probably being overlooked by our method. We make our codes for benchmark, evaluation and method publicly available under \href{https://github.com/mrcoee/Automatic-Label-Error-Detection.git}{GitHub}.

%\MR{TODO: Conclusion and Outlook}

\section*{Acknowledgment}
%We thank \MR{TBA} for discussion and useful advice.
We thank H.\ Gottschalk for discussion and useful advice. We acknowledge support by the European Regional Development Fund (ERDF), grant-no.\ EFRE-0400216. Additionally, this work is funded by the German Federal Ministry for Economic Affairs and Energy, within the project ``KI Delta Learning'', grant no.\ 19A19013Q. We thank the consortium for the successful cooperation. The authors also gratefully acknowledge the Gauss Centre for Supercomputing e.V. (\texttt{https://www.gausscentre.eu}) for funding this project by providing computing time through the John von Neumann Institute for Computing (NIC) on the GCS Supercomputer JUWELS at Julich Supercomputing Centre (JSC). \\ \includegraphics[width=.5\linewidth]{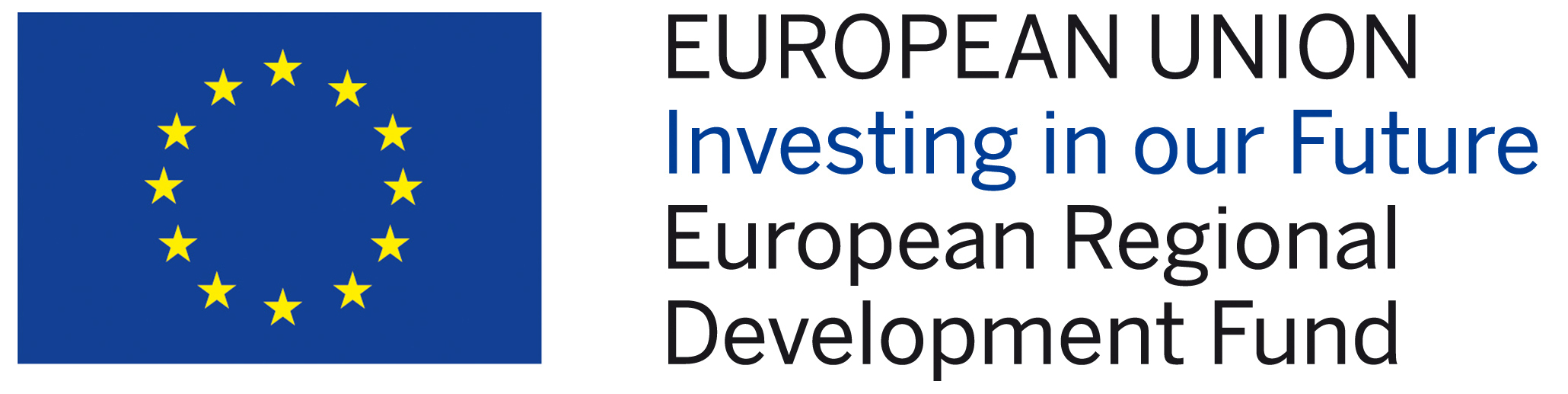}

%\newpage
{\small
\bibliographystyle{ieee_fullname}
\bibliography{egbib}
}
%\newpage

\newpage
\textcolor{white}{x}

\newpage

\appendix
\section{Details on Dataset Creation and the Injection of Label Errors}\label{App:A}
\paragraph{CARLA.}
\begin{figure}[h!]
    \centering
    \begin{tikzpicture}
        \node at (0,-2.5)
        {\includegraphics[width=.47\textwidth]{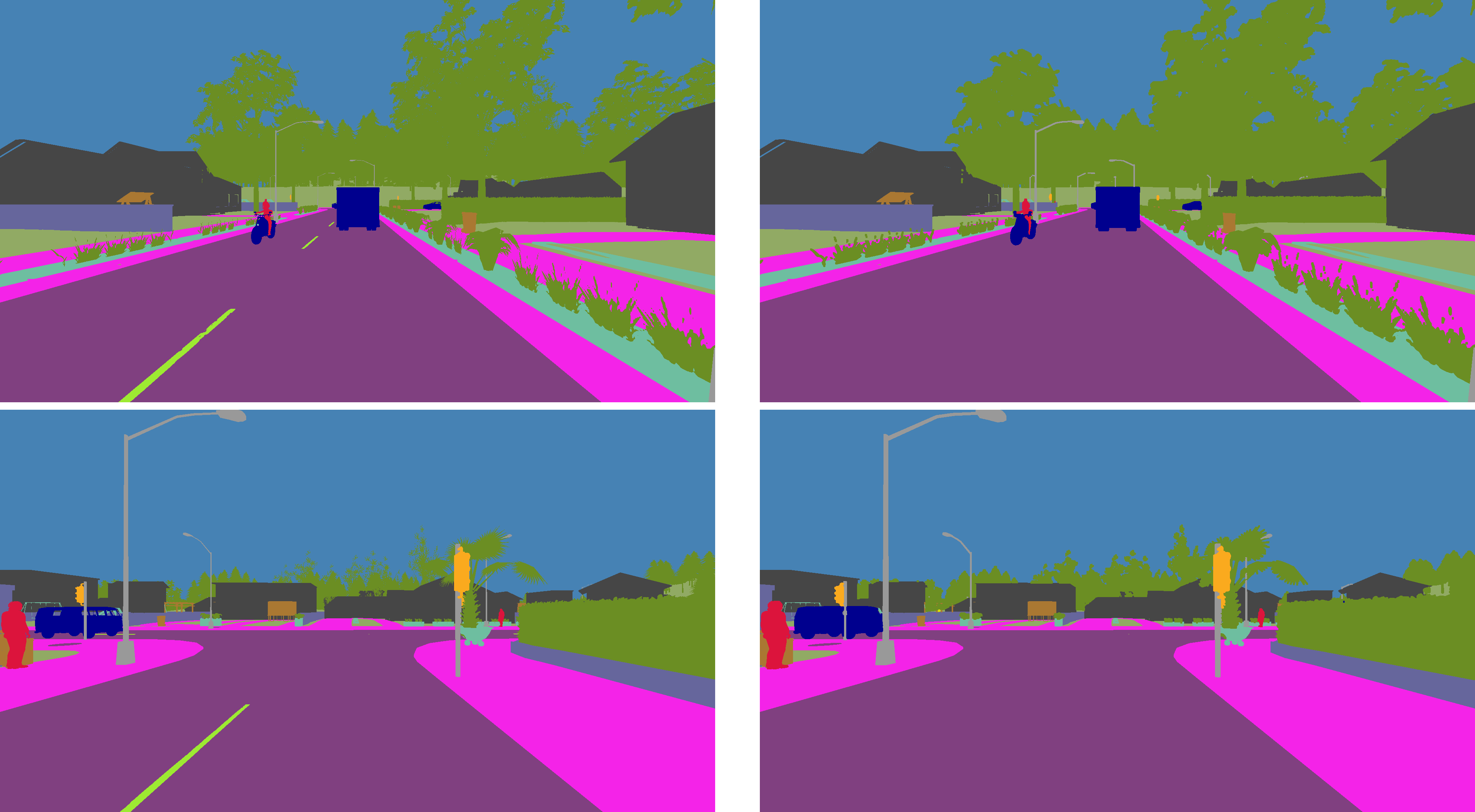}};
        \node at (-2.2, 0.1) {\footnotesize Original Annotation};
        \node at (2.2, 0.1) {\footnotesize Smoothed Annotation};
    \end{tikzpicture}
    \caption{\small Two annotations, before and after we applied smoothing. Note that, we do not smooth every class. Smoothing is applied onto 
    the classes of pedestrians, poles, vegetation and vehicles. Additionally, we remove road lines from the annotation.}
    \label{fig:smoothing}
\end{figure}
We create the dataset by randomly selecting a spawn point of the ego vehicle and recording in the first step a segmentation mask of the empty scene. We then randomly spawn objects around the ego vehicle and record an rgb image as well as another segmentation mask of the scene, while all objects including the ego vehicle are static. The latter pair of image and ground truth serve as input image and unperturbed / clean reference ground truth for evaluation. Thereafter, we copy the connected component of the clean annotation mask into the first one and randomly select 
components to omit in order to induce label errors. When recording the same scene multiple times in CARLA, the ground truth typically varies at boundaries of connected components due to rendering effects. Our procedure of ground truth generation avoids this effect. %By this procedure we eliminate the effect that the label masks provided by CARLA are uncertain on boundaries of connected components due to rendering effects, so that perturbed and unperturbed ground truth are equal except for the induced labeling errors.

The CARLA simulator's semantic segmentation sensor provides very precised annotated segmentation masks. To emulate human annotated masks and to make our synthetic dataset more realistic,
we smooth these masks in order to reduce the degree of detailedness. We proceed class by class and consider the classes of Pedestrians, Poles, Vegetation, and Vehicles (in arbitrary order). 
More precisely, our smoothing process proceeds as follows:
For a given street scene, we first create binary maps for each class where every pixel of the class under consideration for smoothing obtains some value $i \in \mathbb{N}$ and we associate $0$ to every 
other pixel. The value $i$ enables us to control how much an object is smoothed. For our smoothing we set $i=10$. We then apply a Gaussian smoothing kernel onto the binary masks and use these smoothed binary masks as index map to overwrite the values of the corresponding pixels in the original segmentation mask with the currently smoothed class value. 
To avoid that we override values of pixels that belong to a component that is in fact in front (in terms of visual depth) of a component of the smoothed class, we use depth information provided by CARLA to determine during smoothing for each pixel which class is in front. After we finished this smoothing process we fill the windows of vehicles in the scene as CARLA ignores them in the labeling process.
Lastly, we remove the road lines and overwrite them with the class road. The result of this process can be seen in \cref{fig:smoothing}.

\paragraph{Cityscapes.}
Due to the availability of polygonal annotations and the labeling style being hierarchical (objects are mostly labeled on top of the background components), we can drop polygons according to the procedure mentioned in \cref{eq:perturbation_rate}. In this way, we obtain a perturbed ground truth by dropping polygons and then generating the pixel-wise segmentation masks. Although the labeling style is hierarchical, it is possible that the removal of a polygon in the annotation leads to an unlabeled region in the resulting perturbed mask. As it turned out in \cref{sec:numexp}, this is a realistic error and therefore we do not ignore unlabeled components in our evaluation but consider them as a class in the dataset.

\section{A Class-wise Breakdown of Results for Induced Label Errors.}
In this section we present the class-wise results for Cityscapes in experiment $1$ with pertubation rate $\hat{p} = 0.5$ using the sota weights for Nvidia's multiscale attention net (\cref{table:cs_class_nvidia}) and for the 
Deeplab net (\cref{table:cs_class_deeplab}). Ignoring the class Bus, for the attention net we obtained the best $\mathrm{F1}$ score for the class Person with $74.88\%$ which also contains the most label errors. The lowest $\mathrm{F1}$ scores occur for the classes Traffic Light and Traffic Sign. From a visual inspection of predicted masks, it seems that the DNN relies more on the geometry of these objects, while paying less attention to their textures. Similarly, for the Deeplab architecture the class Person achieves the second highest
$\mathrm{F1}$ score with $76.64\%$. Here, the best class is Rider with $80.39\%$ which has a $\mathrm{F1}$ score of only $58.99\%$ with Nvidia's net. The lowest scores are again given for class Traffic Light and class Traffic Sign.

\begin{table}[h]
\label{table:cs_class_nvidia}
\resizebox{0.45\textwidth}{!}{
\begin{tabular}{@{}l@{\hskip 0.3cm}rrrrrrrrr@{}}
\toprule
    & $\mathrm{(m)IoU}$ & $\mathrm{TP}$   & $\mathrm{FN}$   & $\mathrm{FP}$  & $\mathrm{Prec}$ & $\mathrm{Rec}$ & $\mathrm{F1}$\\ \hline\hline
    Bicycle & 84.26 & 81 & 44 & 26 & 75.70 & 64.80 & 69.83\\
    Bus & 95.48 & 1 & 0 & 1 & 50.00  & 100.00 & 66.67\\
    Car & 96.86 & 100 & 61 & 56 & 64.10 & 62.11 & 63.09\\
    Motorcycle & 78.24 & 8 & 6 & 0  & 100.00 & 57.14 & 72.73\\
    Person & 87.91 & 161 & 76 & 32  & 83.42 & 67.93 & 74.88\\
    Rider & 75.47 & 41 & 46 & 11 & 78.85 & 47.13 & 58.99\\
    Traffic Light & 79.88 & 38 & 1 & 63  & 37.62 & 97.44 & 54.29\\
    Traffic Sign & 87.64 & 55 & 17 & 179  & 23.50 & 76.39 & 35.05\\
    Truck & 92.64 & 2 & 8 & 3 & 35.59 & 20.00 & 26.67\\
\midrule
    Overall & 86.82 & 487 & 259 & 371 & 56.76 & 65.28 & 60.72 \\
\midrule
\end{tabular}}
\vspace{.5em}
\caption{Class-wise results for Cityscapes using the Nvidia's Multiscale Attention net.}
\end{table}

\begin{table}[h]
\label{table:cs_class_deeplab}
\resizebox{0.45\textwidth}{!}{
\begin{tabular}{@{}l@{\hskip 0.3cm}rrrrrrrrr@{}}
\toprule
    & $\mathrm{(m)IoU}$ & $\mathrm{TP}$   & $\mathrm{FN}$   & $\mathrm{FP}$ & $\mathrm{Prec}$ & $\mathrm{Rec}$ & $\mathrm{F1}$\\ \hline\hline
    Bicycle & 79.00 & 76 & 49 & 24  & 76.00 & 60.80 & 67.56\\
    Bus & 94.00 & 1 & 0 & 0  & 100.00 & 100.00 & 100.00\\
    Car & 96.50 & 102 & 59 & 29 & 77.86 & 63.35 & 69.86\\
    Motorcycle & 73.80 & 7 & 7 & 1  & 87.50 & 50.00 & 63.64\\
    Person & 88.20 & 159 & 78 & 19  & 89.33 & 67.09 & 76.63\\
    Rider & 75.40 & 43 & 44 & 2  & 95.56 & 49.43 & 80.39\\
    Traffic Light & 79.00 & 33 & 6 & 44  & 62.86 & 84.62 & 56.90\\
    Traffic Sign & 82.80 & 56 & 16 & 206 & 21.37 & 77.78 & 33.53\\
    Truck & 78.80 & 4 & 6 & 0 & 100.00 & 40.00 & 57.14\\
    \midrule
    Overall & 81.40 & 481 & 265 & 325 & 59.68 & 64.48 & 61.98\\
\midrule
\end{tabular}}
\vspace{.5em}
\caption{Class-wise results for Cityscapes using the Deeplab net.}
\end{table}

% \section{Conservative Evaluation of Real Label Errors.} \label{appendix:conservative}
% Due to our conservative validation approach, for the class traffic light we validated several predictions as false positive even though it is very likely that they are actually true positives; see \cref{fig:tls} for one example. 

% \begin{figure}[h!]
%     \center
%     \caption{The segment shown here is originally labeled as void. Our method predicts a label error here which is likely true. However, since we cannot confirm this without any doubt we validated it
%         as false positive.}
%     {\includegraphics[width=.47\textwidth]{../images/tls.png}}
%     \label{fig:tls}
% \end{figure}

% Validating them as true positive would result in additional $15$ true positives and a precision 
% of $93\%$ for the class of traffic lights and of $65\%$ in total. Also, in \cref{fig:fg} we present a case that occurred several times and is somewhat debatable whether it is a false or true positives. However, following the official Cityscapes labeling policy and to avoid any confusion, we consider them to be false positives.

% \begin{figure}[h!]
%     \center
%     \caption{The neural net has correctly found a car segment here. However, it is not a foreground object and therefore labeled as void in the ground truth. According to the labeling, policy this is  
%     a false positive.}
%     {\includegraphics[width=.47\textwidth]{../images/fg.png}}
%     \label{fig:fg}
% \end{figure}

\section{Class-wise Breakdown of Found Label Errors in Frequently used Datasets}\label{app:label_breakdown}
This section is an extension to the discussion of \cref{sec:realerrors}. Here we provide additional details
for each dataset we studied and present results for selected classes. In \cref{sec:label_errors} we present for each dataset a collection of label errors found in the respective datasets.

\paragraph{Cityscapes.}
\begin{figure}[h!]
    \center
    {\includegraphics[width=.47\textwidth]{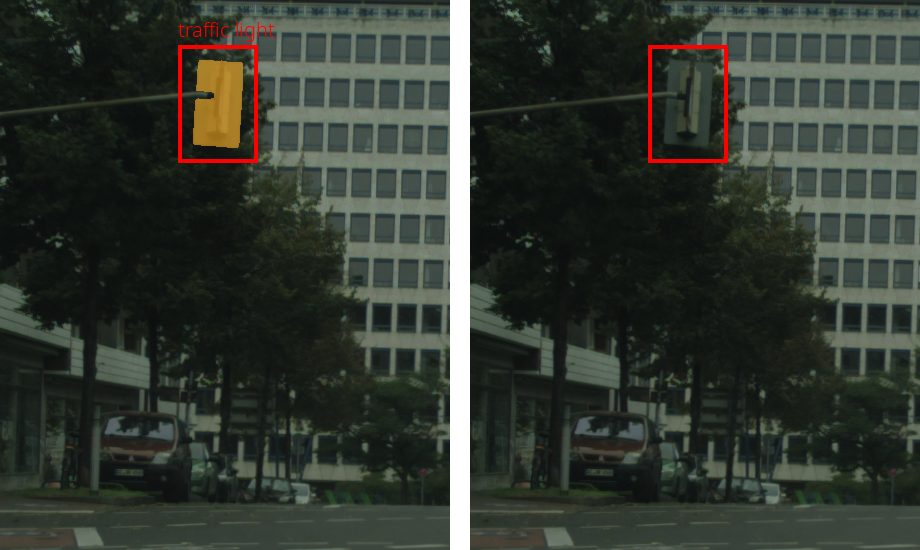}}
    \label{fig:tls}
    \caption{The connected component shown here is originally labeled as void. Our method predicts a label error here which is likely true. However, since we cannot confirm this without any doubt we validated it
        as false positive.}
\end{figure}
In \cref{table:cs_study_val} we have given the class-wise results for Cityscapes and \cref{fig:cs_ex} presents example errors we found. In the training set 
we have found $106$ label errors by reviewing $200$ predictions. For the classes of person, rider/bicycle and vehicles altogether we found $90$ true positives, $60$ false positive, obtaining
a precision of $60\%$. For traffic lights and road signs we achieve a precision of $50\%$ by finding $25$ label errors in $50$ predictions. Apart from the class bus which only consists 
of one prediction, we observe the highest precision for the class rider with $88.89\%$ while, at the same time  the $\mathrm{IoU}$ is the lowest of all predicted classes with $75.47\%$. This indicates 
that our approach does not necessarily depend on a high $\mathrm{IoU}$. It is also possible that the computed $\mathrm{IoU}$ for this class is dragged down by label errors in this class. The most errors we found for this class were caused by labeling a rider as a person. 

We observed the same for the validation set. Combining the results of the classes of person, rider/bicycle and vehicles we have a precision of $55.33\%$ and of $46.00\%$ for classes of traffic lights and road signs.
The result are slightly worse, which is to be expected as the validation set only consists of $500$ images while the training set includes almost $3000$ images.

Due to our conservative validation approach, for the class traffic light we validated several predictions as false positive even though they could be viewed as true positives; see \cref{fig:tls} for examples.

\begin{figure}[h!]
    \center
    {\includegraphics[width=.47\textwidth]{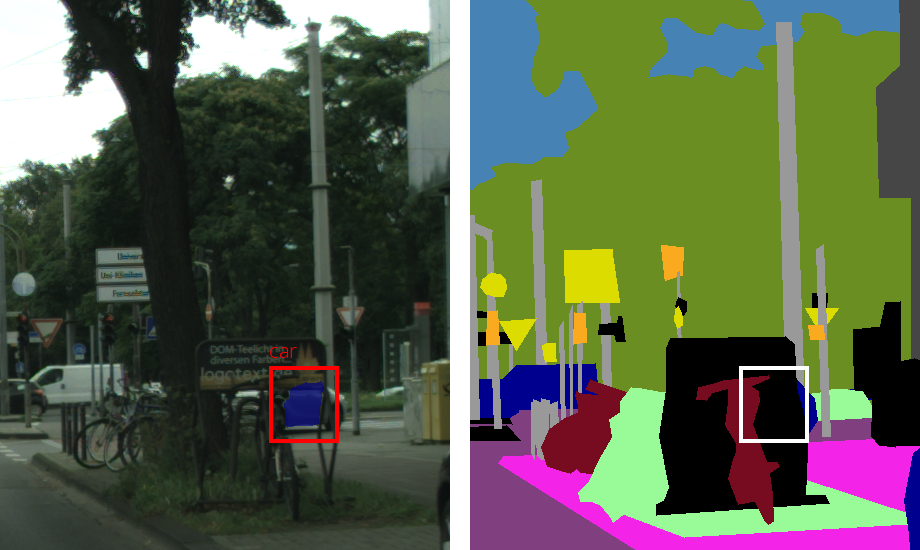}}
    \caption{The DNN has correctly found a car component here. However, it is not a foreground object and therefore labeled as void in the ground truth. According to the labeling policy this is  
    a false positive.}
    \label{fig:fg}
\end{figure}

Validating them as true positive would result in additional $15$ true positives and a precision 
of $93\%$ for the class of traffic lights and of $65\%$ in total. Also, in \cref{fig:fg} we present a case that occurred several times and is debatable whether it is a false or true positives. However, following the official Cityscapes labeling policy and to avoid any confusion, we consider them to be false positives.

\begin{table}[h]
    \label{table:cs_study_val}
    \resizebox{0.45\textwidth}{!}{
    \begin{tabular}{ lrrrr | rrr }
    \toprule
    \multicolumn{5}{c|}{Training set} & \multicolumn{3}{c}{Validation set} \\ \hline
     Class & $\mathrm{(m)IoU}$ & $\mathrm{TP}$ & $\mathrm{FP}$ & $\mathrm{Prec}$ & $\mathrm{TP}$ & $\mathrm{FP}$ & $\mathrm{Prec}$ \\
    \hline\hline
    Bicycle & 84.26 & 16 & 8 & 66.67 & 16 & 8 & 66.67 \\
    Bus & 95.48 & 1 & 0 & 100.00 & 2 & 3 & 40.00 \\
    Car & 96.86 & 13 & 36 & 26.53 & 30 & 32 & 48.39 \\
    Motorcycle & 78.24 & 2 & 1 & 66.67 & 0 & 0 & NaN \\
    Person & 87.91 & 34 & 17 & 66.67 & 35 & 13 & 72.91 \\
    Rider & 75.47 & 16 & 2 & 88.89 & 7 & 2 & 77.78 \\
    Traffic sign & 79.88 & 10 & 18 & 35.71 & 12 & 17 & 41.38 \\
    Traffic light & 87.64 & 13 & 9 & 59.09 & 13 & 8 & 61.90 \\
    Truck & 92.64 & 1 & 3 & 25.00 & 0 & 2 & 00.00 \\
    \midrule
    Overall & 86.82 & 106 & 94 & 53.00 & 115 & 85 & 57.50 \\ 
    \midrule
    \end{tabular}}
    \vspace{.5em}
    \caption{Classwise results for the train (left) and validation (right) sets of Cityscapes.}
\end{table}

\begin{table*}[t!]
    \label{table:real_e_study}
    \resizebox{1.0\textwidth}{!}{
    \begin{tabular}{ lrrrr | lrrrr | lrrrr }
    \toprule
    \multicolumn{5}{c|}{PascalVOC} & \multicolumn{5}{c|}{COCO-Stuff} & \multicolumn{5}{c}{ADE20K} \\  \hline
    Class & $\mathrm{(m)IoU}$ & $\mathrm{TP}$ & $\mathrm{FP}$ & $\mathrm{Prec}$ & Class & $\mathrm{(m)IoU}$ & $\mathrm{TP}$ & $\mathrm{FP}$ & $\mathrm{Prec}$ & Class & $\mathrm{(m)IoU}$ & $\mathrm{TP}$ & $\mathrm{FP}$ & $\mathrm{Prec}$\\
    \hline\hline
    % \hspace*{0.25cm}\footnotesize{\emph{PascalVOC}} & & & & & \hspace*{0.25cm}\footnotesize{\emph{COCO-Stuff}} & & & & & \hspace*{0.25cm}\footnotesize{\emph{ADE20K}} \\
    Bicycle & 78.78 & 0 & 5 & 00.00 & Building-other & 30.81 & 16 & 7 & 69.57 & Building & 83.89 & 0 & 17 & 00.00 \\
    Boat & 66.76 & 4 & 3 & 57.14 & Ceiling-other & 93.76 & 5 & 2 & 71.42 & Cabinet & 65.40 & 4 & 2 & 66.67 \\
    Bottle & 81.18 & 5 & 6 & 45.45 & Desk-stuff & 66.83 & 4 & 0 & 100.00 & Chair & 57.72 & 3 & 3 & 50.00 \\
    Car & 83.64 & 8 & 8 & 50.00 & Grass & 90.81 & 11 & 6 & 64.71 & Coffee & 64.15 & 6 & 0 & 100.00 \\
    Cat & 89.21 & 1 & 10 & 09.09 & Pavement & 26.77 & 3 & 3 & 50.00 & Floor & 80.32 & 28 & 5 & 84.85 \\
    Chair & 51.97 & 17 & 11 & 60.71 & Playingfield & 44.87 & 9 & 0 & 100.00 & Grass & 59.71 & 1 & 5 & 16.67 \\
    Dining table & 51.47 & 10 & 6 & 62.50 & River & 85.00 & 1 & 1 & 50.00 & House & 25.05 & 0 & 3 & 00.00 \\
    Dog & 85.01 & 0 & 14 & 00.00 & Road & 90.70 & 2 & 2 & 50.00 & Painting & 73.64 & 1 & 4 & 20.00 \\
    Person & 85.79 & 38 & 7 & 84.44 & Sea & 96.65 & 3 & 2 & 60.00 & Sea & 47.57 & 5 & 1 & 83.33 \\
    Potted plant & 68.02 & 0 & 4 & 00.00 & Sky-other & 63.13 & 17 & 2 & 89.47 & Ship & 4.58 & 2 & 3 & 40.00 \\
    Sheep & 79.61 & 2 & 3 & 40.00 & Snow & 97.21 & 6 & 2 & 75.00 & Sky & 92.79 & 6 & 3 & 66.67 \\
    Sofa & 52.02 & 6 & 9 & 40.00 & Table & 00.00 & 0 & 6 & 00.00 & Table & 59.24 & 4 & 1 & 80.00 \\
    Train & 82.94 & 1 & 5 & 16.67 & Tree & 73.33 & 31 & 9 & 77.50 & Tree & 73.58 & 3 & 1 & 75.00 \\
    Tv monitor & 68.67 & 0 & 5 & 00.00 & Wall-concrete & 58.59 & 9 & 11 & 45.00 & Wall & 75.24 & 7 & 12 & 36.84 \\
    \midrule
    Overall & 78.03 & 95 & 105 & 47.50 & Overall & 28.20 & 134 & 66 & 67.00 & Overall & 43.12 & 110 & 90 & 55.00 \\ 
    \midrule
    \end{tabular}}
    \vspace{.5em}
    \caption{Classwise results for the validation set of PascalVOC, COCO, and ADE20K.}
\end{table*}

\paragraph{PascalVOC.}
For PascalVOC, COCO, and ADE20K we presented a selection of the most informative classes in \cref{table:real_e_study}. Examples errors are shown in \cref{fig:ps_ex}. 
At first glance, one might expect that classes with low $\mathrm{IoU}$ are difficult for our label error detection. PascalVOC provides several counter example for this.
%a low $\mathrm{IoU}$ would corresponce with a poor result.
While for the class Person we observe high $\mathrm{IoU}$ scores together with strong label error detection performance, considering the overall picture there is a clear anti-correlation between label error detection performance and $\mathrm{IoU}$ for PascalVOC. Even with low IoU scores, our method is still able to find label errors. For example, the DNN exhibits a comparatively low $\mathrm{IoU}$ score of $51.97\%$ for the chair class but still our method found $17$ label errors with a precision of $61\%$. We observe the same for the class Dining table in which we found $10$ label errors with 
a precision of $0.63$ and an $\mathrm{IoU}$ of $51.47$. Calculating the Pearson correlation between the IoU and the precision scores we obtain a low negative value of $-0.27$.
Altogether, the findings indicate that the low $\mathrm{mIoU}$ might be partially caused by a poor label quality of the dataset.

\paragraph{COCO.}
Selected class results are given in \cref{table:real_e_study} and some errors are presented in \cref{fig:coco_ex}.
% Some classes appear to be underrepresented in the dataset \MR{MR: das hier klingt gemutmasst, aber dazu sollte es doch eine Statistik geben. Hier sollte man nicht ueber ein Erscheinungsbild argumentieren}, therefore it is unclear how meaningful these $\mathrm{IoU}$ scores are. 
The best results are achieved for the classes of 
Building-other, Grass, Playingfield, Sky-other and Tree with precision scores significantly above $50\%$ and a representative amount of predictions. Overall, for this dataset we achieved 
the highest precision of $67\%$ in $200$ prediction across $36$ predicted classes. 
For this dataset we also can observe the IoU and precision scores also seem to be barley correlated. This observation is supported by a 
low correlation score of $0.36$.
% \MR{No additional interesting findings?}

\paragraph{ADE20K.}
In this dataset we found $104$ label errors in $200$ predictions across $58$ classes. Table \ref{table:real_e_study} shows results for selected classes and \cref{fig:ade_ex} some example errors. As we already mentioned 
the class definitions of this dataset are in part not sufficiently distinct. In addition, since we were not able to find class descriptions for this dataset, we inferred from the ground truth annotation which objects the classes represent. 
This led to a significant amount of false positives since we were oftentimes unable to assign $\mathrm{TP}$ with appropriate confidence.
This applies in particular to the class ``building''. In ADE20K, there also exists a class named house which seems to accommodate buildings for human habitation. For the DNN it is difficult to distinguish these two classes, hence it predicts most of the buildings/houses as buildings 
% \MR{MR: is building also the predominant label?}. 
This leads to a high $\mathrm{IoU}$ for the class building, a low $\mathrm{IoU}$ of $25.05$ for the class house, and a precision of $0\%$ for the building class. However, despite the mentioned issues, the results for ADE20K are comparable to those obtained for the other datasets. Also for this dataset we have a low correlation score of $0.19$.

%\newpage

\section{Additional Examples of label Errors in Frequently used Datasets}\label{sec:label_errors}
\begin{figure*}[t]
    \center
    {\includegraphics[width=.88\textwidth]{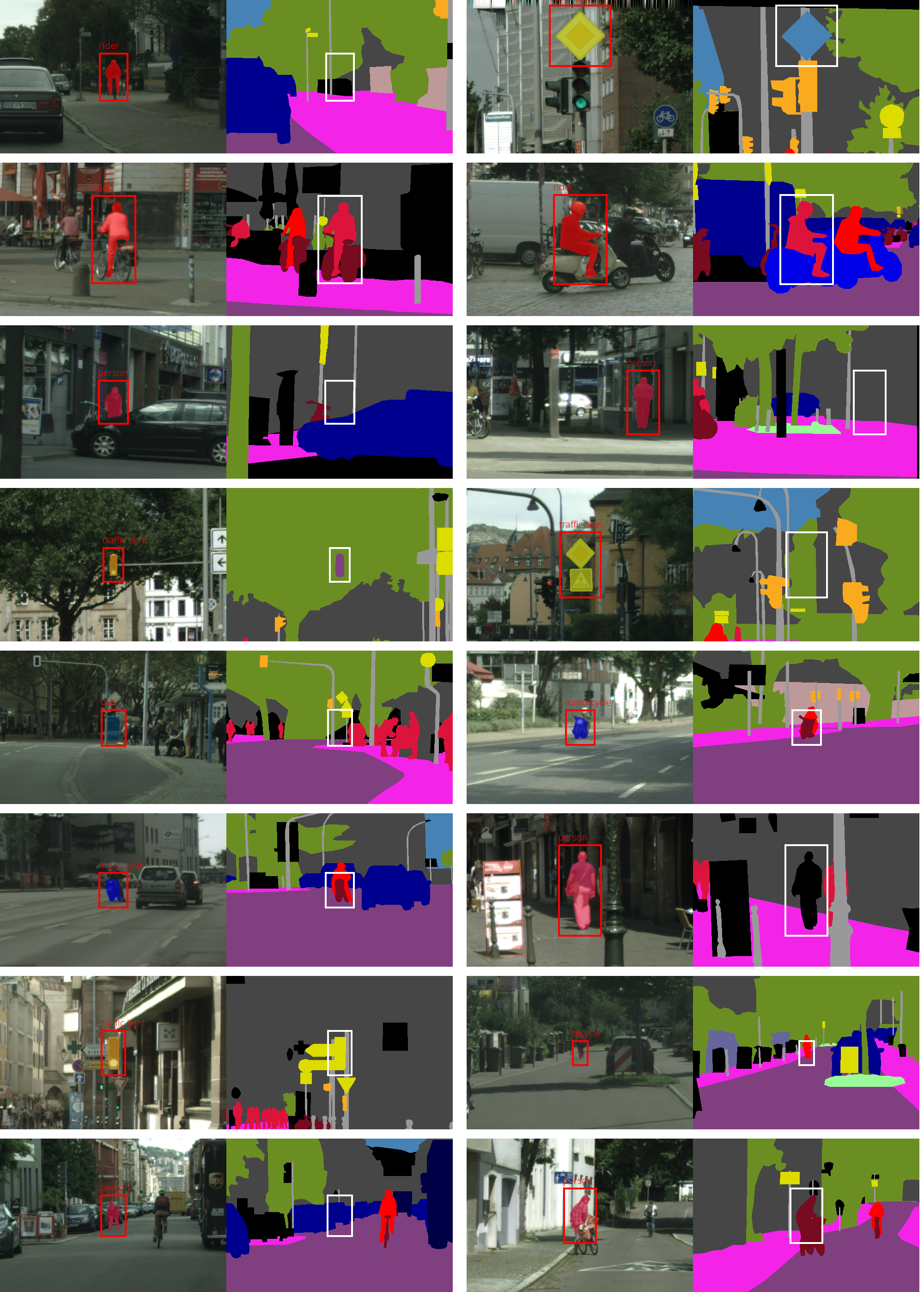}}
    \caption{A collection of label errors present in Cityscapes. Left: prediction of our label error detection method; right: ``ground truth'' annotation. Our method is able to find both, overlooked and flipped labels.}
    \label{fig:cs_ex}
\end{figure*}

\begin{figure*}[t]
    \center
    {\includegraphics[width=.88\textwidth]{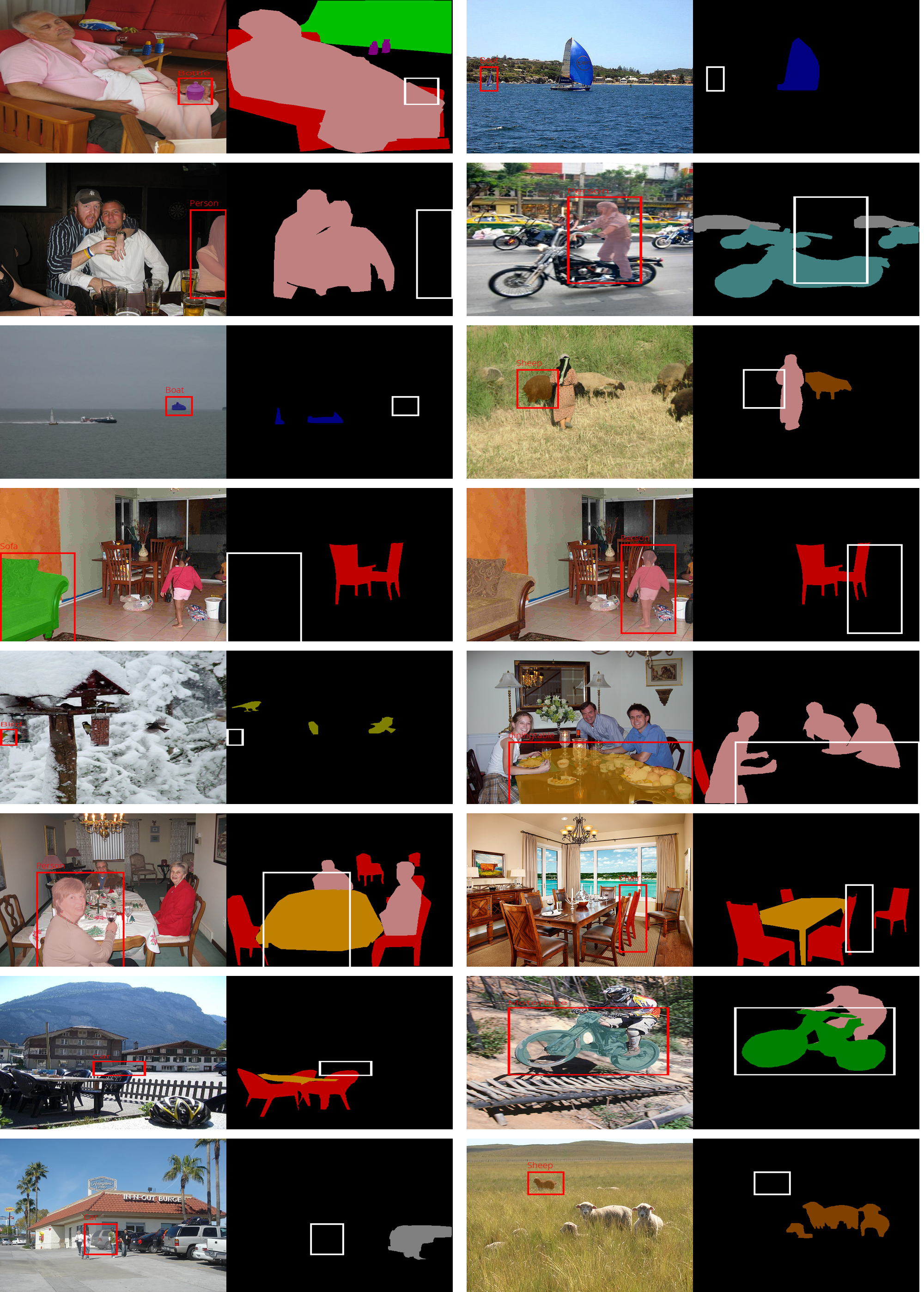}}
    \caption{A collection of label errors present in PascalVOC. The visualization scheme follows the one of \cref{fig:cs_ex}.}
    \label{fig:ps_ex}
\end{figure*}

\begin{figure*}[t]
    \center
    {\includegraphics[width=.88\textwidth]{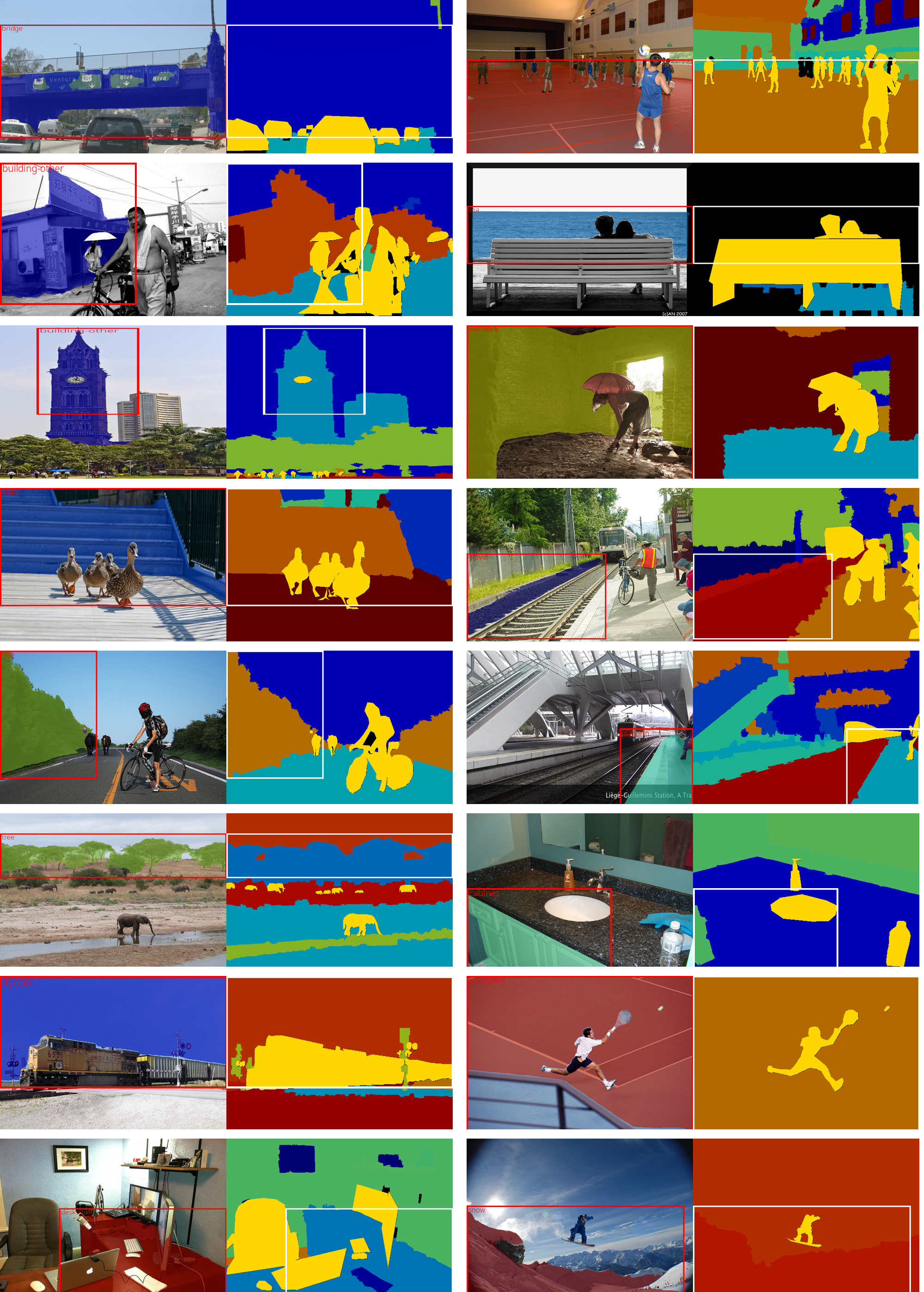}}
    \caption{A collection of label errors present in COCO-Stuff. The visualization scheme follows the one of \cref{fig:cs_ex}.}
    \label{fig:coco_ex}
\end{figure*}

\begin{figure*}[t]
    \center
    {\includegraphics[width=.88\textwidth]{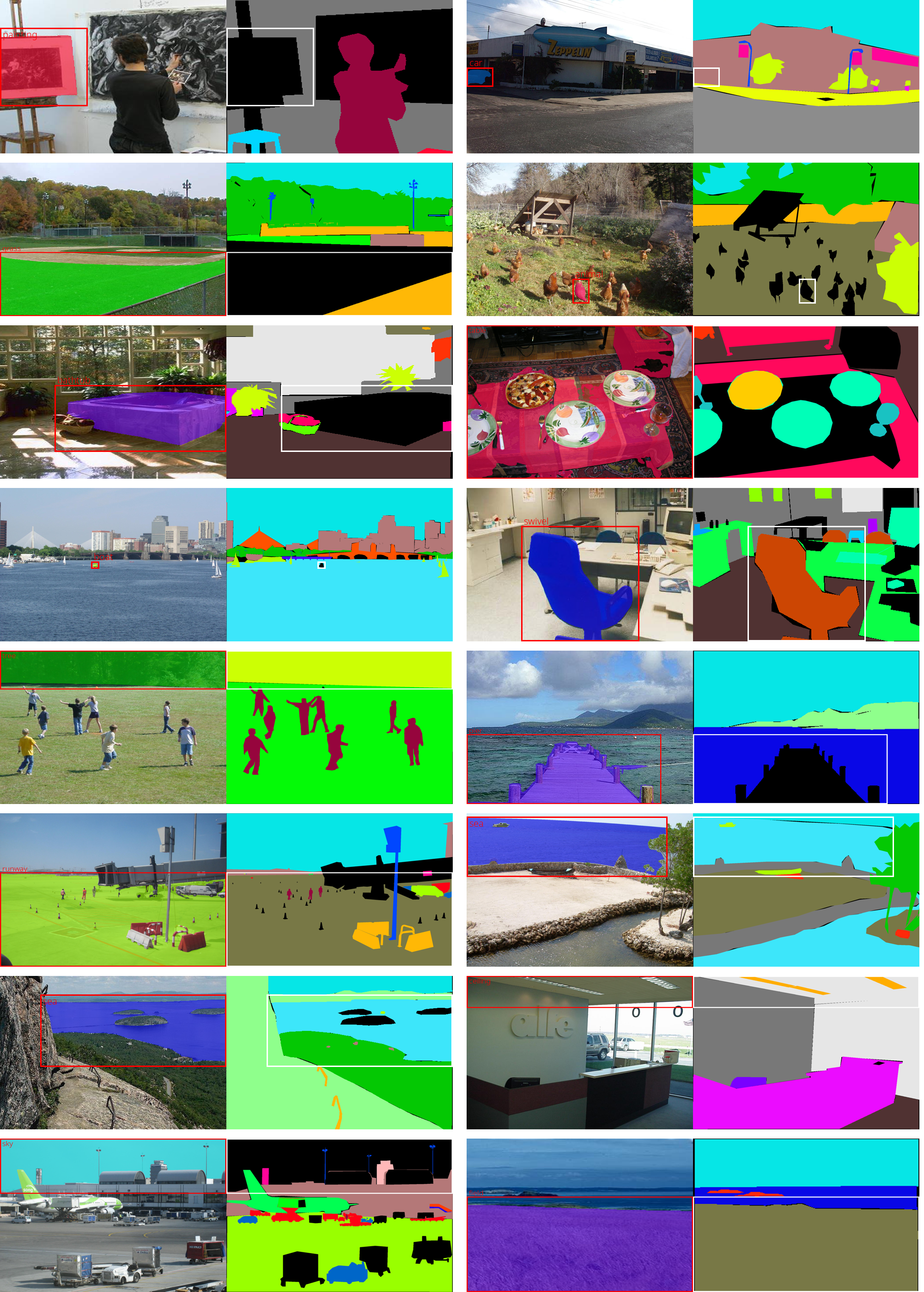}}
    \caption{A collection of label errors present in ADE20K. The visualization scheme follows the one of \cref{fig:cs_ex}.}
    \label{fig:ade_ex}
\end{figure*}

Below we present collages of examplary label errors we found in the datasets discussed in the previous section. Each collage contains $16$ pairs of images where each pair represent one label error. The right image represents a section of the ground truth segmentation mask and the left image
displays a missed or flipped component. \Cref{fig:cs_ex,fig:ps_ex,fig:coco_ex,fig:ade_ex} are collections of label errors in Cityscapes, PascalVOC, COCO-Stuff and ADE20K, respectively.

\end{document}